\documentclass[12pt]{article}
\usepackage{amsmath,amssymb,amsthm, cite,mathrsfs,color,authblk}
\usepackage[colorlinks, linkcolor=blue]{hyperref}

\usepackage{graphicx}
\usepackage{subfigure}

\usepackage{enumitem}

\usepackage{booktabs, threeparttable}

\usepackage{tabularx}

\usepackage{multirow, multicol}

\usepackage{geometry}
\usepackage{titlesec}
\titleformat{\section}
  {\normalfont\fontsize{14}{12}\bfseries}{\thesection}{1em}{}

\titleformat{\subsection}
  {\normalfont\fontsize{12}{12}\bfseries}{\thesubsection}{1em}{}

\begin{document}

%==============Title, Author, Abstract==================%
\title{ A Deep Neural Network Framework for Solving Forward and Inverse Problems in Delay Differential Equations}

\author[a]{Housen Wang}
\author[a]{Yuxing Chen}
\author[a]{Sirong Cao}
\author[a]{Xiaoli Wang}
\author[a]{Qiang Liu\thanks{Corresponding author: Qiang Liu, ~Email address: matliu@szu.edu.cn.
\\
This work partially supported by research grants: Guangdong Basic and Applied Basic Research Foundation (2022B1515120009), the Science and Technology Program of Shenzhen (20231121110406001)}}
\affil[a]{School of Mathematical Sciences, Shenzhen University, Shenzhen, 518060, China}
\renewcommand*{\Affilfont}{\small\it}
\date{}

\maketitle

\begin{abstract}
We propose a unified framework for delay differential equations (DDEs) based on deep neural networks (DNNs) - the neural delay differential equations (NDDEs), aimed at solving the forward and inverse problems of delay differential equations. This framework could embed delay differential equations into neural networks to accommodate the diverse requirements of DDEs in terms of initial conditions, control equations, and known data. NDDEs adjust the network parameters through automatic differentiation and optimization algorithms to minimize the loss function, thereby obtaining numerical solutions to the delay differential equations without the grid dependence and polynomial interpolation typical of traditional numerical methods. In addressing inverse problems, the NDDE framework can utilize observational data to perform precise estimation of single or multiple delay parameters, which is very important in practical mathematical modeling. The results of multiple numerical experiments have shown that NDDEs demonstrate high precision in both forward and inverse problems, proving their effectiveness and promising potential in dealing with delayed differential equation issues.
\end{abstract}

%============== Introduction ==================%
% \begin{multicols}{2}
\section{Introduction}
Delay differential equations (DDEs) have broad applications in many important fields such as population dynamics\cite{hutchinson1948circular}, fluid dynamics\cite{villermaux1994pulsed}, ecology\cite{BANKS201714}, biology\cite{Glass2021}, and medicine\cite{CULSHAW200027}. We refer the reader to the book \cite{erneux2009applied} for a fairly large application and its references therein. However, computing the numerical solutions of DDEs is a very challenging task. The delay terms cause the current state to depend not only on the variables' values at the current time but also on the values at one or more past time, which greatly increase computational complexity. Finite difference methods and asymptotic solution methods are typically employed to solve DDEs. Finite difference methods\cite{bellen1985numerical, in1996stability, maset2000stability, 2019Impulsive, bellen2013numerical} discretize the time interval and the spatial domain using grid partitioning, storing or calculating the values of the solution at past time points on discrete time points. They then transform the continuous DDEs into a series of algebraic or difference equations, finally obtaining expressions for approximate values of the solution at each time point, thus achieving numerical approximate solutions over the entire time interval. It is difficult to handle the delayed values in the numerical scheme since we need to evaluate the intermediate values triggered by delays. In general, a polynomial interpolation between two finite nodes is necessary.
Asymptotic solution methods\cite{cooke1994numerical, gulsu2012new, hazarika2024application, luo2024class} simplify complex DDEs into a series of approximate problems to obtain solutions to the original problem. Despite the progress made by these methods in DDEs, there are challenges in accurately handling delay terms, managing historical data, and maintaining long-term stability. Moreover, the efficiency and effectiveness of these methods are limited when solving DDEs with multiple delays or time-varying delays. 
On the other hand, estimating model parameters is crucial for modeling using delayed differential equations. To confront such problems, people have used approaches such as least squares, the maximum likelihood method, Bayesian analysis and so on \cite{rosen1984discrete, bocharov1994numerical, moreno1994regulation, baker1997modelling, mehrkanoon2014parameter, yang2024inverse}.
At this point, it is generally needed to design a new algorithm performed by minimization of an objective function.

Deep learning has been widely applied in numerous areas such as image classification\cite{krizhevsky2017imagenet}, speech processing\cite{mehrish2023review}, natural language processing\cite{khan2023exploring}, and recommendation systems\cite{zhang2019deep}, etc. 
In recent years, deep learning models have achieved significant results in solving differential equation problems with the growth of computational resources. 
The physics informed neural networks\cite{RAISSI2019686} embeds physical information in the framework as prior knowledge, achieving great success in dealing with partial differential equations. 
By approximating complicated functions effectively, artificial neural networks show promising application prospects in solving numerical solutions and estimating invariant parameters of dynamical systems controlled by differential equations \cite{yuan2022pinn, srati2024physics}. 
Compared to traditional methods, solutions obtained by neural networks for solving differential equations problems are continuous and differentiable, ensuring higher accuracy without the need for discretization. 
Nevertheless, neural networks for solving the inverse problem of DDEs that include both delay and system parameters have been much less explored.

In this article, we propose a novel deep neural network framework - Neural Delay Differential Equations (NDDEs) to obtain numerical solutions to DDEs and estimate their parameters. In the NDDEs framework, the input layer of deep neural networks (DNNs) is responsible for receiving initial data or system parameters. The hidden layers perform complex nonlinear transformations, and the output layer provides numerical solutions. Each neuron in the network is connected to the previous layer via weights and utilizes activation functions to achieve nonlinear transformations, enabling the network to capture correlations among complex functions. By designing appropriate loss functions, we embed the relationships of delay differential equations into the network structure and use optimizers to adjust parameters, thereby obtaining numerical solutions to the equations.

The proposed network method is used to solve both the forward and inverse problems of delay differential equations. The framework will demonstrate great performance on the forward problem of delay differential equation and its system based on given initial conditions. In addition, NDDEs will precisely identify the delays and system parameters in the inverse problem by adjusting neural network parameters through optimization algorithms based on the given data.

The remainder of this article is organized as follows. In Section \ref{sec.Definition}, we provide the definition of delay differential equations and the system of delay differential equations. In Section \ref{sec.NDDEs_Framework}, we propose the NDDEs framework for solving delay differential equations. In Section \ref{sec.Numerical_experiments}, we use the designed NDDEs framework to solve both the forward and the inverse problems of delay differential equations to demonstrate its effectiveness.

%============== Problem Statement ==================%
\section{The Definition of Delay Differential Equations}
\label{sec.Definition}
In real world, any system almost certainly involves time delays, because a finite time is always required to sense information and then respond. A DDE for such model with a single delay is generally of the form
\begin{equation}
    y'(t) = f(t,y(t),y(t-\tau)),~\mathrm{for}\; t \geq \xi,
\label{eq.DDE}
\end{equation}
where $f(t, u, v)$ is a given continuous function, $\xi>0$ is a constant, and $\tau > 0$ is the time delay. An initial condition for Eq. \eqref{eq.DDE} is given by
\begin{equation}
    y(t) = \phi(t),~\mathrm{for}\; t \in [\xi-\tau,\xi],
\label{eq.DDE_IC}
\end{equation}
where $\phi(\cdot)$ is a given continuous function.

The system of DDEs is in the following form
\begin{equation}
    \begin{cases}
        \begin{aligned}
            y_1'(t) ={} & f_1(t,y_1(t), \cdots, y_n(t),y_1(t-\tau_{1,1}), \cdots, y_n(t-\tau_{n,1})) \\
            y_2'(t) ={} & f_2(t,y_1(t), \cdots, y_n(t),y_1(t-\tau_{1,2}), \cdots, y_n(t-\tau_{n,2})) \\
            \vdots ~&\\
            y_n'(t) ={} & f_n(t,y_1(t), \cdots, y_n(t),y_1(t-\tau_{1,n}), \cdots, y_n(t-\tau_{n,n}))
        \end{aligned}
    \end{cases}
\label{eq.DDEs}
\end{equation}
where $f_j(\cdot)$ represents the $j$-th continuous function, and $\tau_{k,j} > 0$ is the time delay parameter for $y_k(\cdot)$ in the $j$-th equation. The initial conditions for Eq. \eqref{eq.DDEs} are given by
\begin{equation}
    y_j(t) = \phi_j(t),~\mathrm{for}\; t \in [\xi-\tau,\xi],\; j=1, 2, \cdots, n,
\label{eq.DDEs_IC}
\end{equation}
where $\phi_j(\cdot)$ is the j-th given continuous function for $y_j(\cdot)$.

Similar to the well-posedness of ordinary differential equations, existence and uniqueness theorem for
the problem \eqref{eq.DDE}-\eqref{eq.DDE_IC} is essentially based on the continuity of the function $f$ and its Lipschitz continuity with respect to $u$ and $v$. While DDEs also have distinct characteristics, such as oscillatory and chaotic behavior, derivative discontinuities, etc \cite{bellen2013numerical}. In general, there are more complex properties for DDEs than ordinary differential equations Whether in theory or numerical analysis.

%============ The Framework of NDDEs ===============%
\section{The Framework of NDDEs}
\label{sec.NDDEs_Framework}

\subsection{The Framework for Solving Delay Differential Equations}
\label{subsec.DDE_Framework_FP}
\begin{figure}[htbp]
    \centering
    \includegraphics[width=0.9\textwidth]{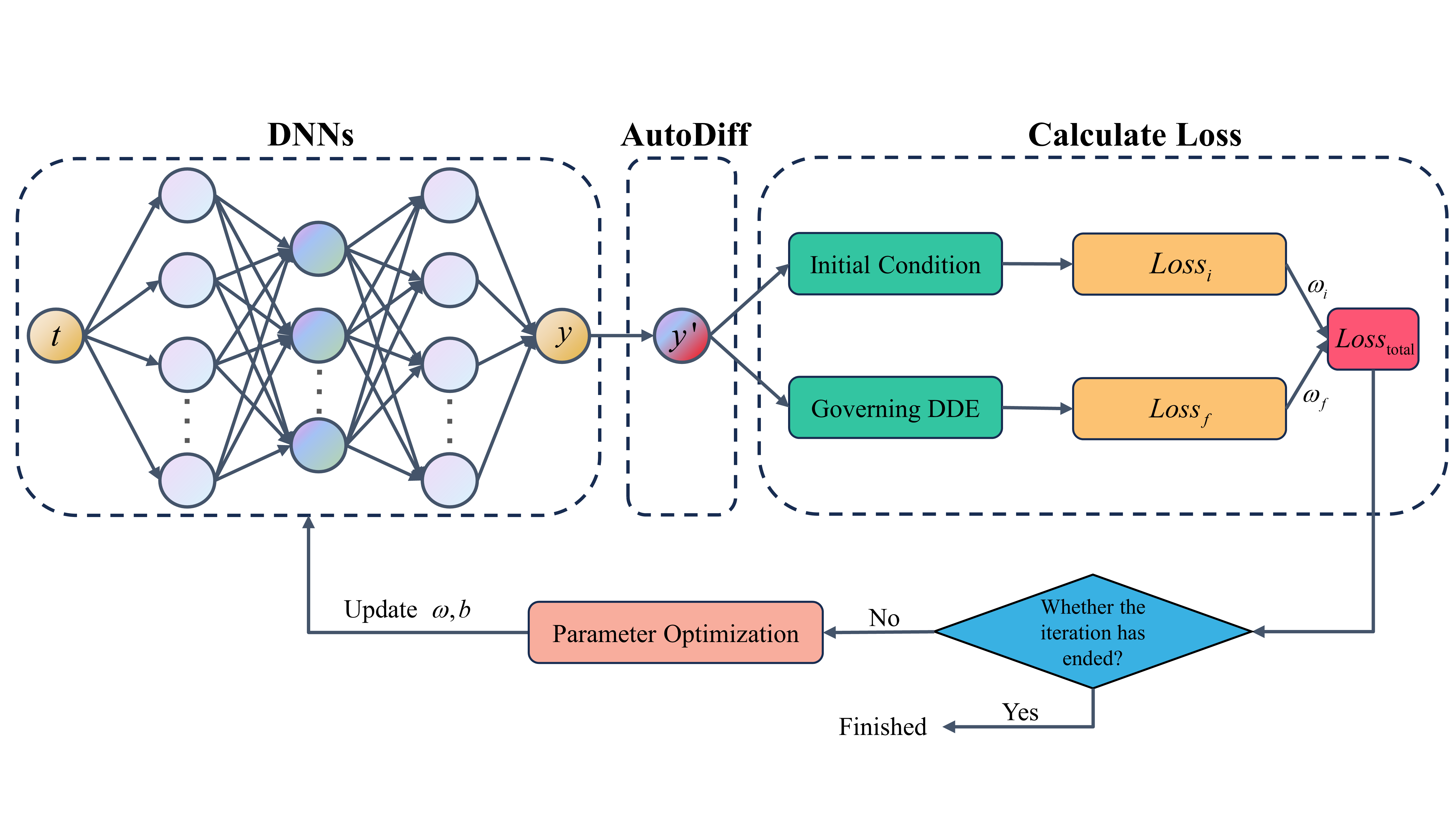} 
    \caption{The framework of NDDEs for solving forward problem of delay differential equations.}
    \label{fig.DDE_FP_framework}
\end{figure}

\begin{figure}[htbp]
    \centering
    \includegraphics[width=0.9\textwidth]{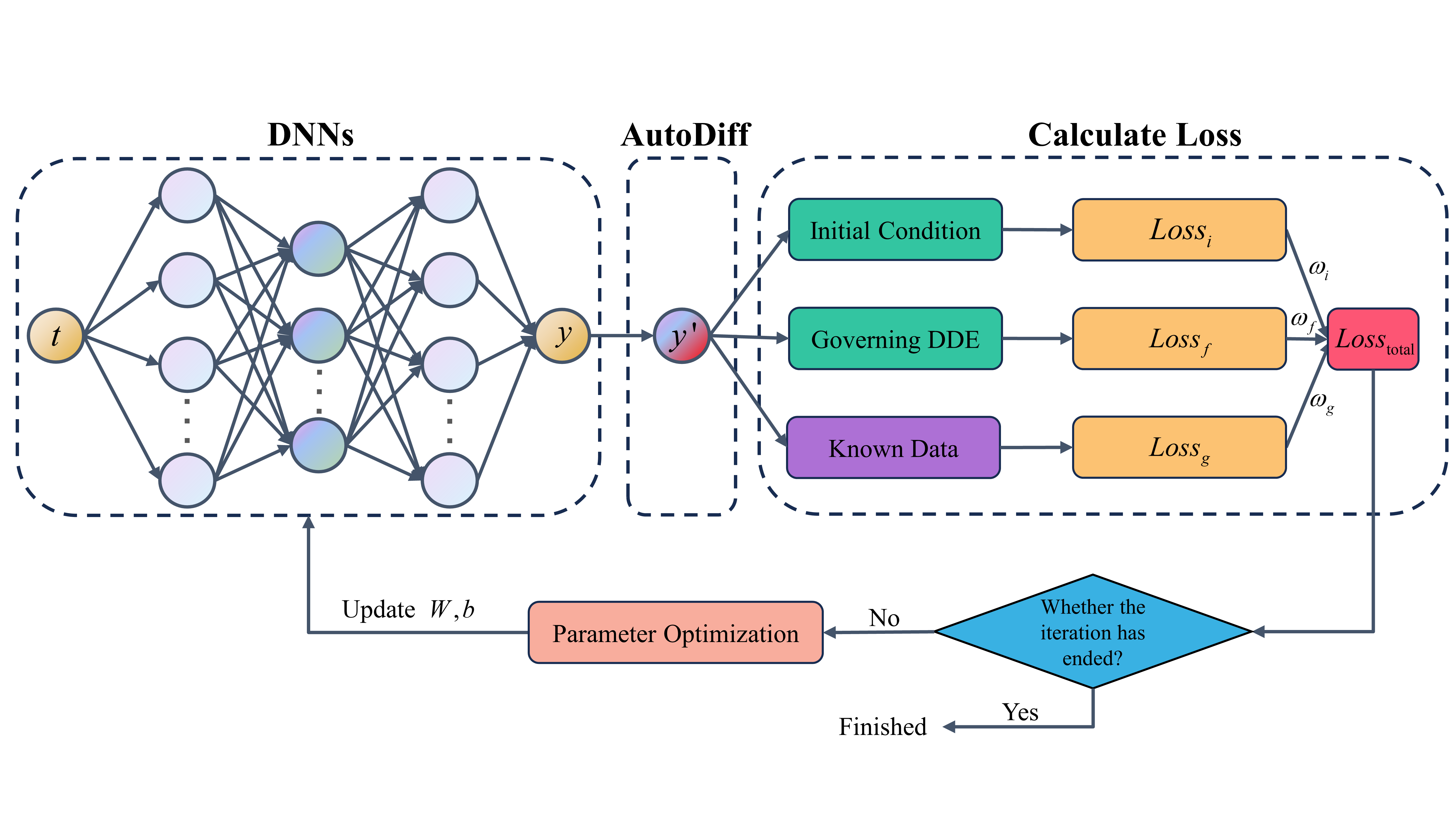} 
    \caption{The framework of NDDEs for solving inverse problem of delay differential equations.}
    \label{fig.DDE_IP_framework}
\end{figure}

This subsection introduces the design of NDDEs to solve the forward and inverse problems of delay differential equations. The framework for solving the forward problem of delay differential equations is shown in Fig. \ref{fig.DDE_FP_framework}, while the framework for the inverse problem is depicted in Fig. \ref{fig.DDE_IP_framework}. Comparing the two methods, we found that the difference between them is not significant. We just add the undetermined parameters to the network for training. Therefore, this framework is unified and does not require too many additional changes.

In general, we consider the delay differential equation given by Eq. \eqref{eq.DDE} with $\xi=0$ in the solution interval $[0,T]$ (where $T>0$):
\begin{equation}
    y'(t) = f(t,y(t),y(t-\tau)),~\mathrm{for}\; t \geq 0,
\label{eq.DDE_FP_framework}
\end{equation}
where $f(t, u, v)$ is a given continuous function and $\tau > 0$ is the time delay. An initial condition for Eq. \eqref{eq.DDE_FP_framework} is given by
\begin{equation}
    y(t) = \phi_0(t), ~\text{for $t \in [-\tau,0]$},
\label{eq.DDE_FP_framework_IC}
\end{equation}
where $\phi_0(\cdot)$ is a given continuous function.

NDDEs approximate the true values $y_{\mathrm{exact}}(t_i)$ with predicted values $y_{\mathrm{pred}}(t_i)$ calculated by DNNs, thus we can obtain the loss on the delay differential equation:
$$
    Loss_f = \dfrac{1}{N_f}\sum_{i=1}^{N_f}\left|\frac{\mathrm{d}}{\mathrm{d}t}y_{\mathrm{pred}}(t_i)- f(t_i,y_{\mathrm{pred}}(t_i),y_{\mathrm{pred}}(t_i-\tau))\right|^2,
%\label{eq.DDE_FP_framework_loss_f}
$$
where $N_f$ is the number of points randomly sampled within the solution domain $[0, T]$ of the equation. 
Since a dealy equation might have derivative discontinuity at $t=0$, the loss of the initial condition is only calculated by
$$
    Loss_i = \left|y_{\mathrm{pred}}(0) - \phi_0(0)\right|^2.
$$

In practical problem solving, it may happen that one part of the loss converges to zero while the other part does not completely converge. To address this problem, we calculate adaptive weights for all losses in each iteration as \cite{yuan2022pinn}, taking the weighted sum of all losses as the total error $Loss_{\mathrm{total}}$. By normalizing, we assign greater weights to larger losses and smaller weights to smaller losses, which accelerates the convergence of the parts with larger losses. This ensures that all parts of the loss converge to zero simultaneously, thus improving the precision and efficiency of the solution. Taken together, the total loss of the forward problem can be expressed as:
\begin{equation}
    Loss_{\mathrm{total}} = \omega_f \cdot Loss_f + \omega_i \cdot Loss_i,
\label{eq.DDE_FP_framework_loss_total}
\end{equation}
where
\begin{equation}
\omega _f=\dfrac{Loss_f}{Loss_f + Loss_i}, \qquad  \omega _i=\dfrac{Loss_i}{Loss_f + Loss_i}.
\label{eq.DDE_FP_framework_weight}
\end{equation}

Compared to the forward problem, NDDEs incorporate the loss of the given data when solving the inverse problem:
$$
    Loss_g = \dfrac{1}{N_g}\sum_{i=1}^{N_g}\left|y_{\mathrm{pred}}(t_i)-y_{\mathrm{exact}}(t_i)\right|^2,
$$
where $N_g$ denotes the number of given data points. Hence, the total loss in the inverse problem can be calculated as
$$
    Loss_{\mathrm{total}} = \omega_f \cdot Loss_f + \omega_i \cdot Loss_i + \omega_g \cdot Loss_g.
$$
The weight for each loss is determined by the following formula:
\begin{equation}
[\omega_f, \omega_i, \omega_g] = \dfrac{[Loss_f, Loss_i, Loss_g]}{Loss_f + Loss_i + Loss_g}.
\label{eq.DDE_IP_framework_weight}
\end{equation}
Whether for the forward problem or the inverse problem, in each iteration, NDDEs calculate the adaptive weights of all losses, thereby getting the weighted sum of losses. Then, the optimizer optimizes the parameters of the neural network to minimize $Loss_\mathrm{total}$, thus obtaining the convergent solution of the DDE.

\subsection{The Framework for Solving the System of Delay Differential Equations}
\label{subsec.DDEs_Framework_FP}
\begin{figure}[htbp]
    \centering
    \includegraphics[width=0.9\textwidth]{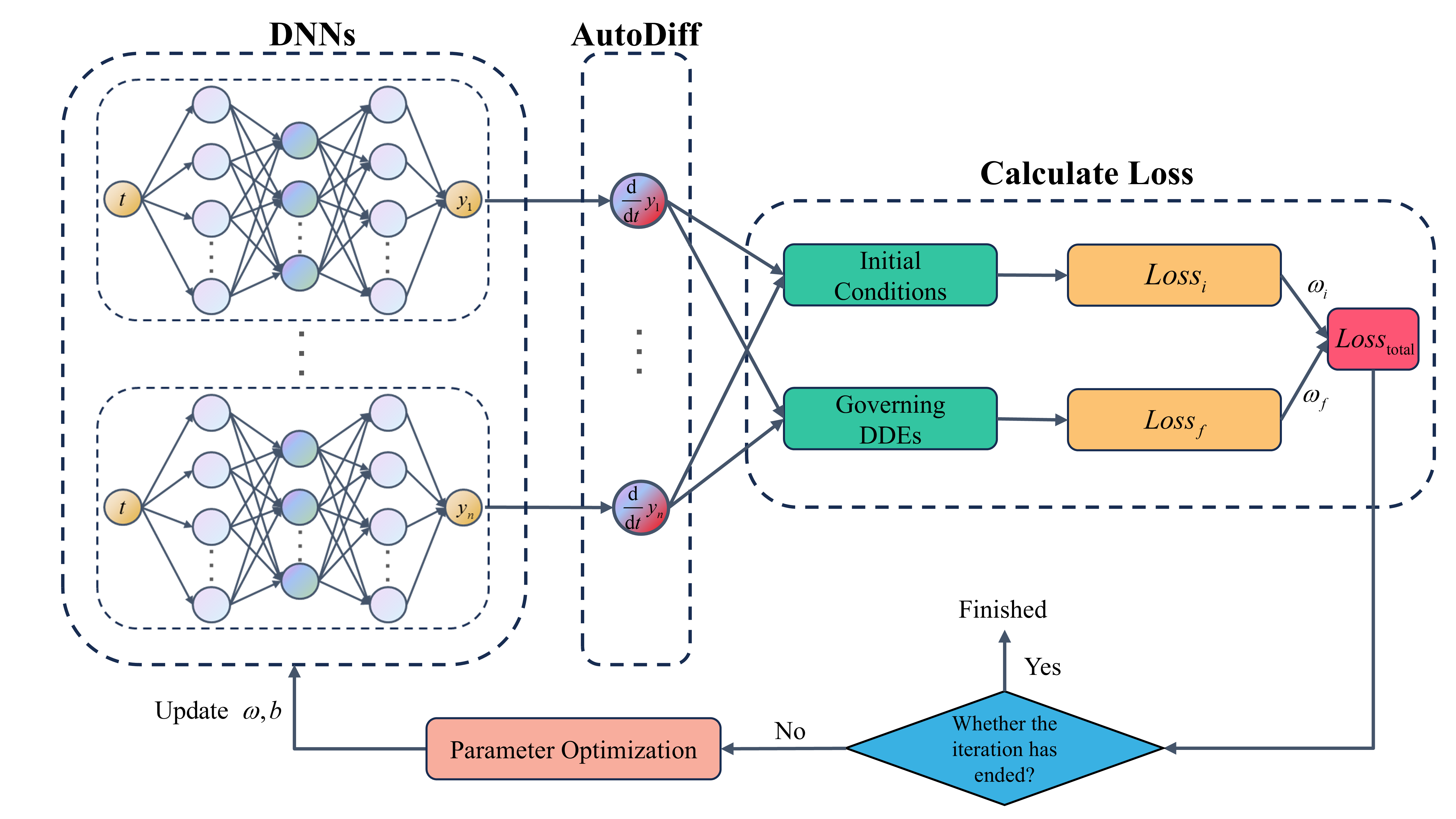} 
    \caption{The framework of NDDEs for solving forward problem of the system of delay differential equations.}
    \label{fig.DDEs_FP_framework}
\end{figure}

This subsection introduces the design of the NDDEs for solving the forward and inverse problems of the system of delay differential equations. By making subtle adjustments to the NDDEs framework for solving DDEs, we can easily deal with the system of delay differential equations. As shown in Fig. \ref{fig.DDEs_FP_framework}, the framework has changed from using a single neural network to using multiple neural networks. Each neural network solves a delay differential equation and the derivative of each output is computed by automatic differentiation.

Consider the system of delay differential equations given by Eq. \eqref{eq.DDEs} with $\xi=0$ in the solution interval $[0,T]$ (where $T>0$):
\begin{equation}
    \begin{cases}
        \begin{aligned}
            y_1'(t) ={} & f_1(t,y_1(t), \cdots, y_n(t),y_1(t-\tau_{1,1}), \cdots, y_n(t-\tau_{n,1})) \\
            y_2'(t) ={} & f_2(t,y_1(t), \cdots, y_n(t),y_1(t-\tau_{1,2}), \cdots, y_n(t-\tau_{n,2})) \\
            \vdots ~&\\
            y_n'(t) ={} & f_n(t,y_1(t), \cdots, y_n(t),y_1(t-\tau_{1,n}), \cdots, y_n(t-\tau_{n,n}))
        \end{aligned}
    \end{cases}
\label{eq.DDEs_FP_framework}
\end{equation}
where $f_j(\cdot)$ represents the $j$-th continuous function, and $\tau_{k,j} > 0$ is the time delay parameter for $y_k(\cdot)$ in the $j$-th equation. The initial conditions for Eq. \eqref{eq.DDEs_FP_framework} are given by
$$
    y_j(t) = \phi_j(t),~\mathrm{for}\; t \in [-\tau,0],\; j=1,2, \cdots, n,
$$
where $\phi_j(\cdot)$ is the $j$-th given continuous function for $y_j(\cdot)$.

By approximating the true values $y^{\mathrm{exact}}_j(t_i)$ with the predicted values $y^{\mathrm{pred}}_j(t_i)$ computed through the DNNs, we can obtain the loss for each equation within the system of delay differential equations:
\begin{align}
    &Loss_{f,j} = \notag 
    \\
    &\dfrac{1}{N_f}\sum_{k=1}^{N_f}\left|\frac{\mathrm{d}}{\mathrm{d}t}y^{\mathrm{pred}}_j(t_k)- f_j\left(t,y^{\mathrm{pred}}_1(t_k), \cdots, y^{\mathrm{pred}}_n(t_k),y^{\mathrm{pred}}_1(t_k-\tau_{1,j}), \cdots, y^{\mathrm{pred}}_n(t-\tau_{n,j})\right)\right|^2,
\label{eq.DDEs_FP_framework_loss_f}
\end{align}
where $N_f$ is the number of points randomly sampled within the solution domain $[0, T]$ of the equation. The Loss of the initial conditions are calculated by
\begin{equation}
    Loss_{i,j} = \left|y^{\mathrm{pred}}_j(0)-\phi_j(0)\right|^2.
\label{eq.DDEs_FP_framework_loss_i}
\end{equation}
Thus, the total loss for the forward problem of the system of delay differential equations can be expressed as:
\begin{align}
    Loss_{\mathrm{total}} 
    = \sum_{j=1}^{n} \left( \omega_{f,j} \cdot Loss_{f,j} + \omega_{i,j} \cdot Loss_{i,j} \right).
\label{eq.DDEs_FP_framework_loss_total}
\end{align}
The weights of each Loss are determined by
\begin{equation}
    \omega_{f,j} = \dfrac{Loss_{f,j}}{\sum\limits_{j=1}^{n} \left( Loss_{f,j} + Loss_{i,j} \right)},
\label{eq.DDEs_FP_weight_f}
\end{equation}
and
\begin{equation}
    \omega_{i,j} = \dfrac{Loss_{i,j}}{\sum\limits_{j=1}^{n} \left( Loss_{f,j} + Loss_{i,j} \right)}.
\label{eq.DDEs_FP_weight_i}
\end{equation}

\begin{figure}[htbp]
    \centering
    \includegraphics[width=0.9\textwidth]{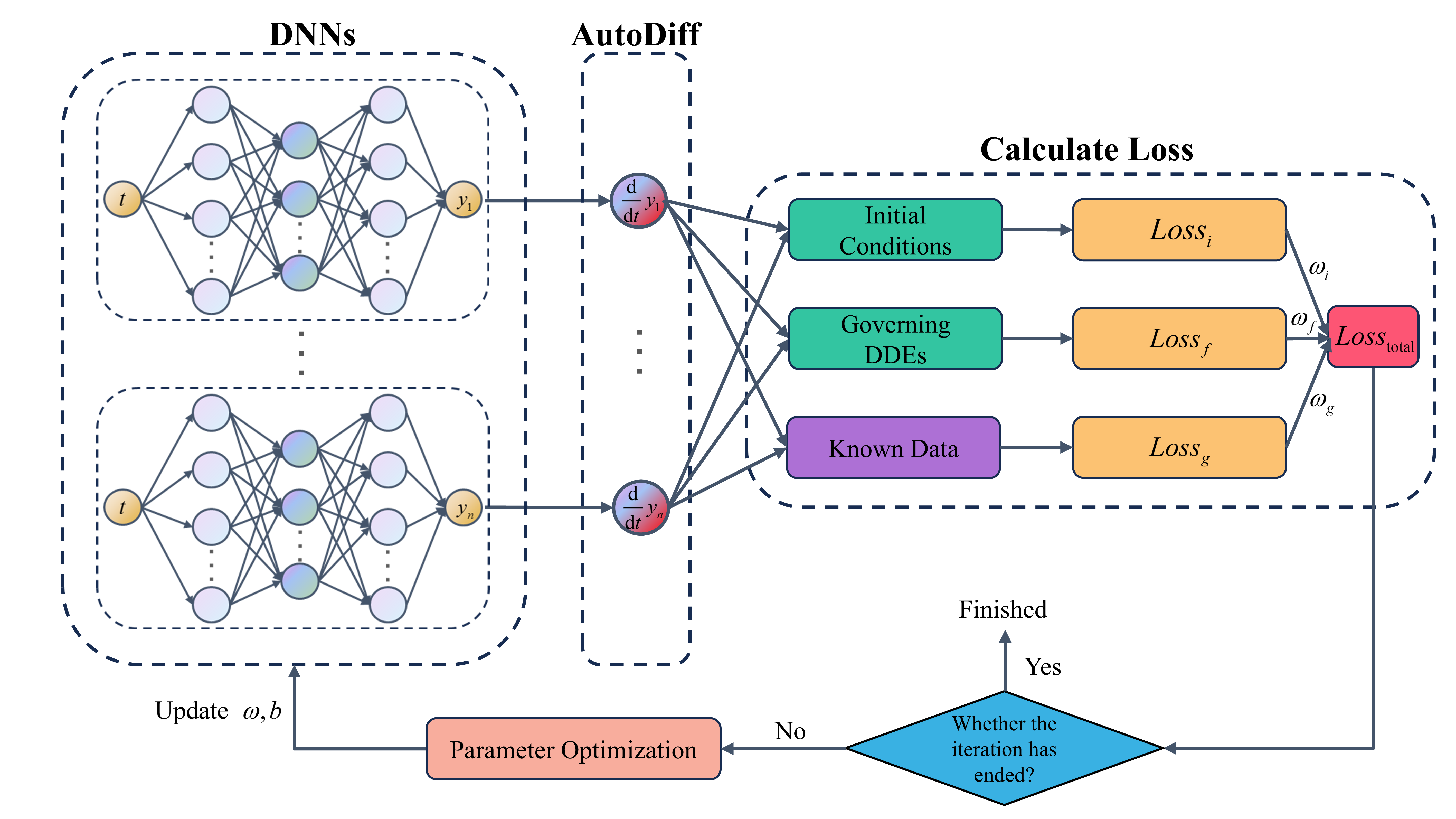} 
    \caption{The framework of NDDEs for solving inverse problem of the system of delay differential equations.}
    \label{fig.DDEs_IP_framework}
\end{figure}

As shown in Fig. \ref{fig.DDEs_IP_framework}, the inverse problem framework complements the forward problem framework by incorporating a loss term for the known data into the loss function. The loss of the given data can be calculated as follows:
\begin{equation}
    Loss_{g,j} = \dfrac{1}{N_g}\sum_{k=1}^{N_g}\left|y^{\mathrm{pred}}_j(t_k)-y^{\mathrm{exact}}_j(t_k)\right|^2,
\label{eq.DDEs_IP_framework_loss_g}
\end{equation}
where $N_g$ denotes the number of given data points. Hence, the total loss in the inverse problem can be calculated as
$$
    Loss_{\mathrm{total}} = \sum_{j=1}^{n} \left( \omega_{f,j} \cdot Loss_{f,j} + \omega_{i,j} \cdot Loss_{i,j} + \omega_{g,j} \cdot Loss_{g,j} \right),
$$
wherein the weights of the respective loss terms are determined by
$$
[\omega_{f,j}, \omega_{i,j}, \omega_{g,j}] = \dfrac{\left[Loss_{f,j},~~ Loss_{i,j},~~ Loss_{g,j}\right]}{\sum\limits_{j=1}^{n} \left( Loss_{f,j} + Loss_{i,j} +  Loss_{g,j} \right)}.
$$

%============ Numerical Experiments ===============%
\section{Numerical Experiments}
\label{sec.Numerical_experiments}

In this section, we will perform numerical experiments to test the performance of the NDDE framework designed in Section \ref{sec.NDDEs_Framework} in solving various delay differential equations and their systems. Specifically, we will assess the solution accuracy of the NDDEs framework for forward problems and evaluate its capability to learn delay and system parameters in inverse problems.  All algorithms are implemented in Python using JAX, an open-source machine learning framework developed by researchers at Google, and the corresponding codes are made available on GitHub \href{code}{https://github.com/HousenWang/NDDEs}.

\subsection{The Forward Problem for Solving Delay Differential Equations}
Consider the delay differential equation with a single time delay
\begin{equation}\label{equ17a}
\begin{cases}
        y'(t) = {} -y(t-\tau),\qquad  & t \geq 0\\
        y(t) = {} 1, & t \leq 0
\end{cases}
\end{equation}
in the interval $[0,10]$. 

We conduct three sets of numerical experiments with the delay parameter $\tau$ respectively taking values of $0.5$, $1.0$, and $1.5$. Taking $\tau=1$ as an example, the loss on the delay differential equation in the NDDEs framework can be expressed as
$$
    Loss_f = \dfrac{1}{N_f}\sum_{i=1}^{N_f}\left|y'_{\mathrm{pred}}(t_i) + y_{\mathrm{pred}}(t_i-1)\right|^2,
$$
the loss of the initial condition is calculated by
$$
    Loss_i = \left|y_{\mathrm{pred}}(0)-1\right|^2.
$$
The total loss is the weighted sum of all the associated losses, which is expressed as in Eq.\eqref{eq.DDE_FP_framework_loss_total}:
\begin{equation}
    Loss_{\mathrm{total}} = \omega_f \cdot Loss_f + \omega_i \cdot Loss_i,
\label{eq.DDE_FP_NE_ex1_loss_total}
\end{equation}
The weights of the losses are determined by the adaptive weighting strategy described in Eq. \eqref{eq.DDE_FP_framework_weight}.

During the numerical solution process, the number of random sampling points $N_f$ in the solution domain is 5,000. The DNNs network structure consists of three hidden layers, with the
number of neurons in each layer being $20$, $40$, and $20$ respectively. We also choose the $\mathrm{tanh}(\cdot)$ as the activation function and use the Adam optimizer to optimize the loss function, with the training conducted 80,000 times.

According to Terpstra's study\cite{terpstra2016delay}, we clearly know that the delay differential equation shown in Eq. \eqref{equ17a} has an explicit analytical solution, which can be specifically expressed as

\begin{equation}  
    y(t) = 1 + \sum_{k=1}^{n} (-1)^k \dfrac{\left[t - (k-1)\tau \right]^k}{k!}, \quad t \in [(n-1)\tau, n\tau], \quad n \in \mathbb{N}.
\label{eq.DDE_FP_ex1_exact}
\end{equation}

To evaluate the precision of the solution, we chose the relative $L^2$ error between the numerical solution from the NDDEs framework and the exact solution as the evaluation criterion. The expression of the relative $L^2$ error is as shown in 
Eq. \eqref{eq.DDE_NE_L2error}, and detailed comparison results is presented in Tab. \ref{tab.DDE_L2error}.

\begin{equation}
\mathrm{The~Relative}~L^2~\mathrm{Error} = \dfrac{\sqrt{\sum_{k=1}^{N_f} \left|y_{\mathrm{pred}}(t_k) - y_{\mathrm{exact}}(t_k)\right|^2}}{\sqrt{\sum_{k=1}^{N_f} \left|y_{\mathrm{exact}}(t_k)\right|^2}}.
\label{eq.DDE_NE_L2error}
\end{equation}

\begin{table}[htbp]
  \centering
  \caption{The relative $L^2$ error between the predicted and exact solution}
    \begin{tabularx}{1\textwidth}{>{\centering\arraybackslash}X>{\centering\arraybackslash}X>{\centering\arraybackslash}X>{\centering\arraybackslash}X}
    \toprule
    Delay Parameter $\tau$   & 0.5 & 1.0 & 1.5 \\
    \midrule
    The $L^2$ RE & 0.1023\% & 0.3349\% & 0.2635\% \\
    \bottomrule
    \end{tabularx}
  \label{tab.DDE_L2error}
\end{table}

\begin{figure}[!htp]
	\begin{minipage}{0.48\textwidth}
		\centering
		\includegraphics[height=5cm]{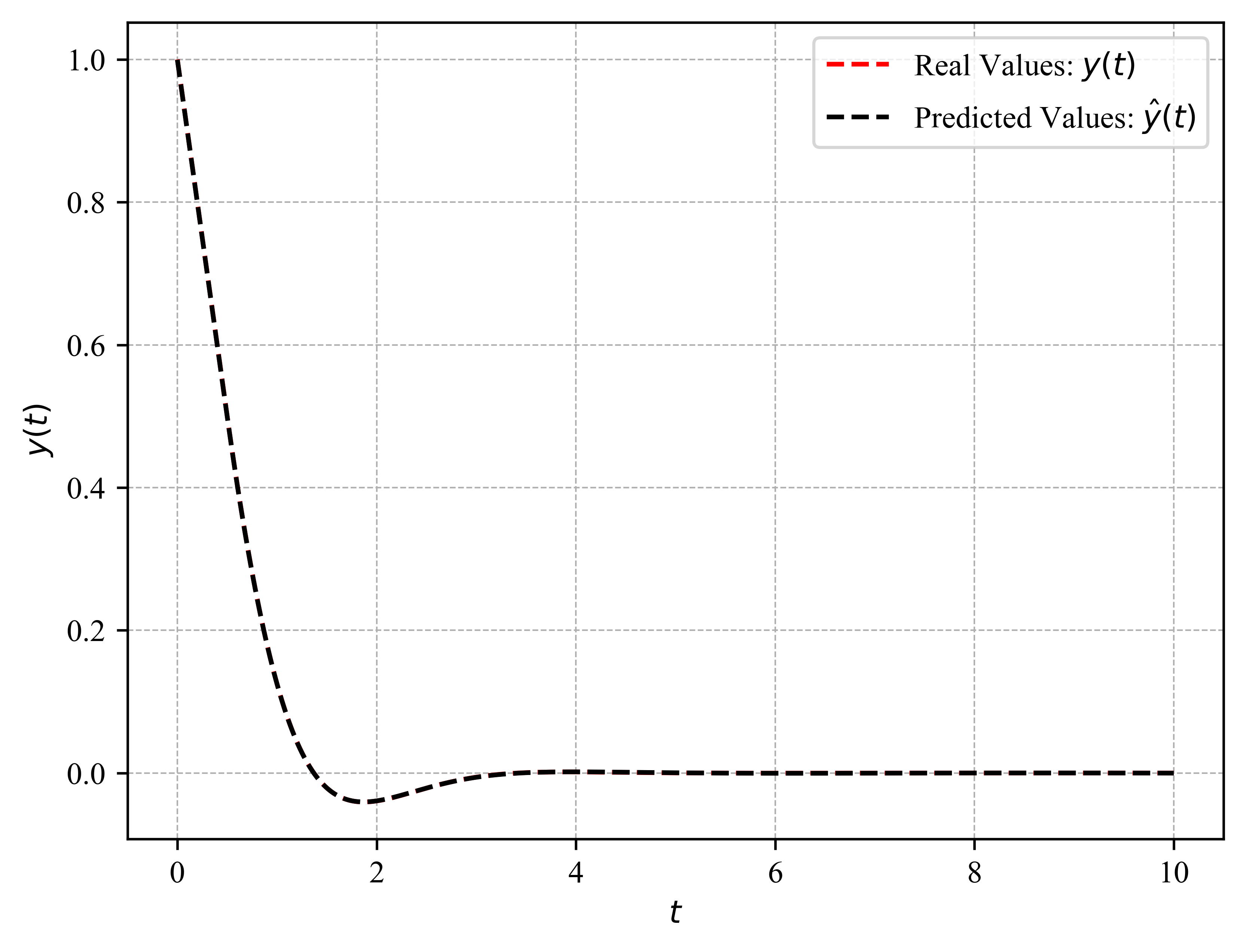}
		\caption{\footnotesize Trajectories of the final predicted vs. true trajectories for the delay differential equation with $\tau=0.5$ as shown in Eq. \eqref{equ17a}.}
		\label{fig:DDE_sol1}
	\end{minipage}\hfill
	\begin{minipage}{0.48\textwidth}
		\centering
		\includegraphics[height=5cm]{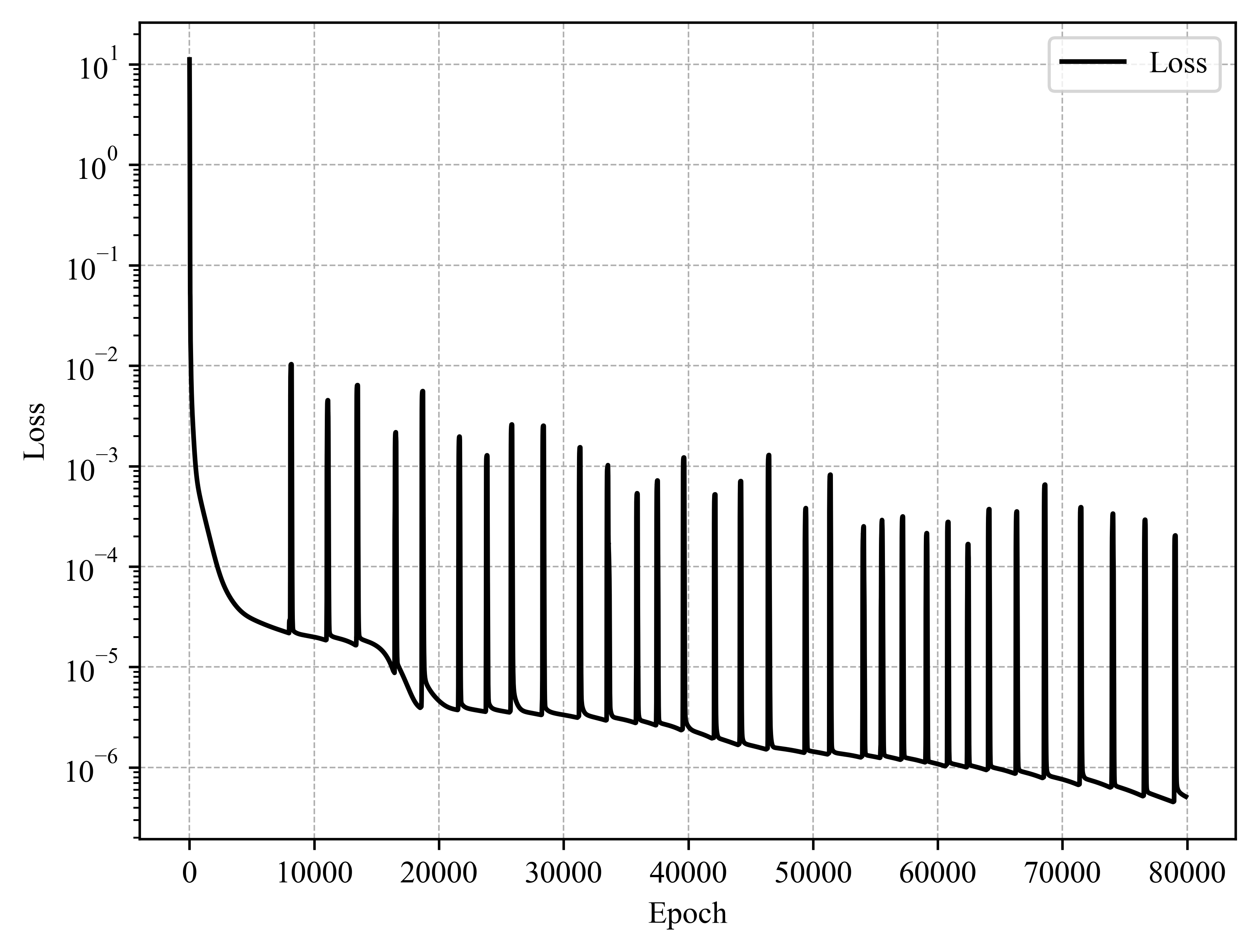}
		\caption{\footnotesize Loss convergence over iterations in NDDEs training when $\tau=0.5$.}
		\label{fig:DDE_loss1}
	\end{minipage}

 	\begin{minipage}{0.48\textwidth}
		\centering
		\includegraphics[height=5cm]{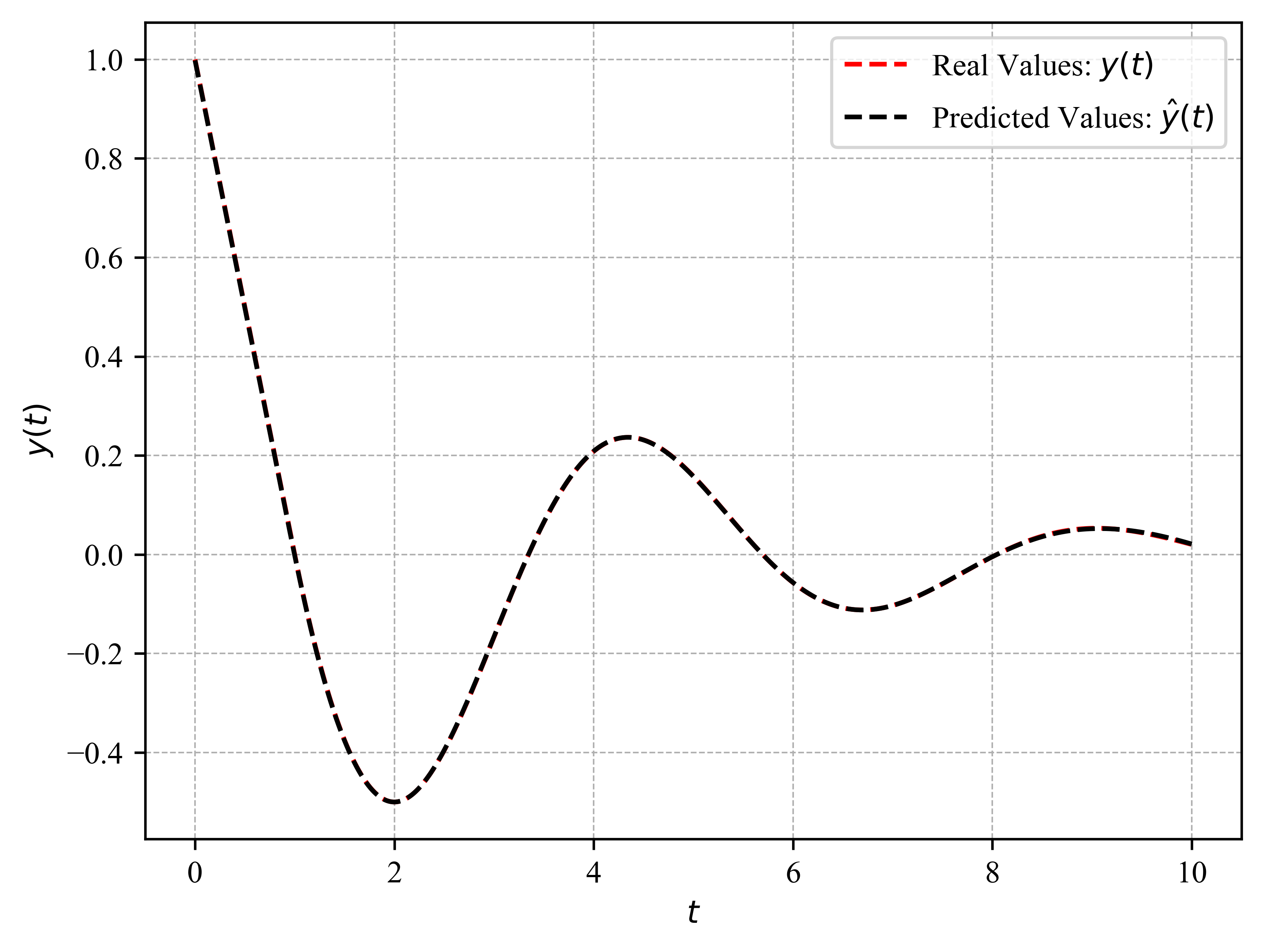}
		\caption{\footnotesize Trajectories of the final predicted vs. true trajectories for the delay differential equation with $\tau=1.0$ as shown in Eq. \eqref{equ17a}.}
		\label{fig:DDE_sol2}
	\end{minipage}\hfill
	\begin{minipage}{0.48\textwidth}
		\centering
		\includegraphics[height=5cm]{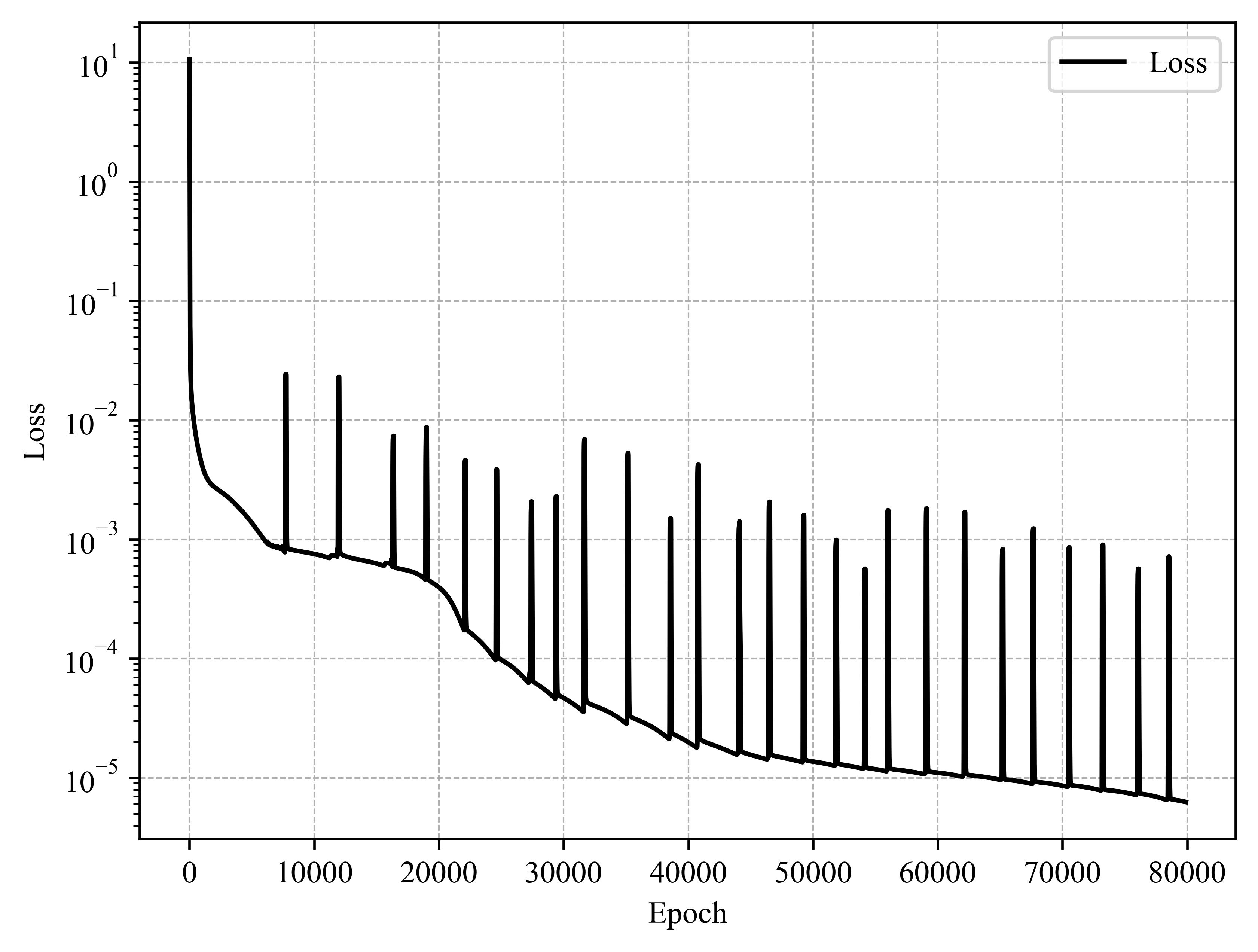}
		\caption{\footnotesize Loss convergence over iterations in NDDEs training when $\tau=1.0$.}
		\label{fig:DDE_loss2}
	\end{minipage}

        \begin{minipage}{0.48\textwidth}
		\centering
		\includegraphics[height=5cm]{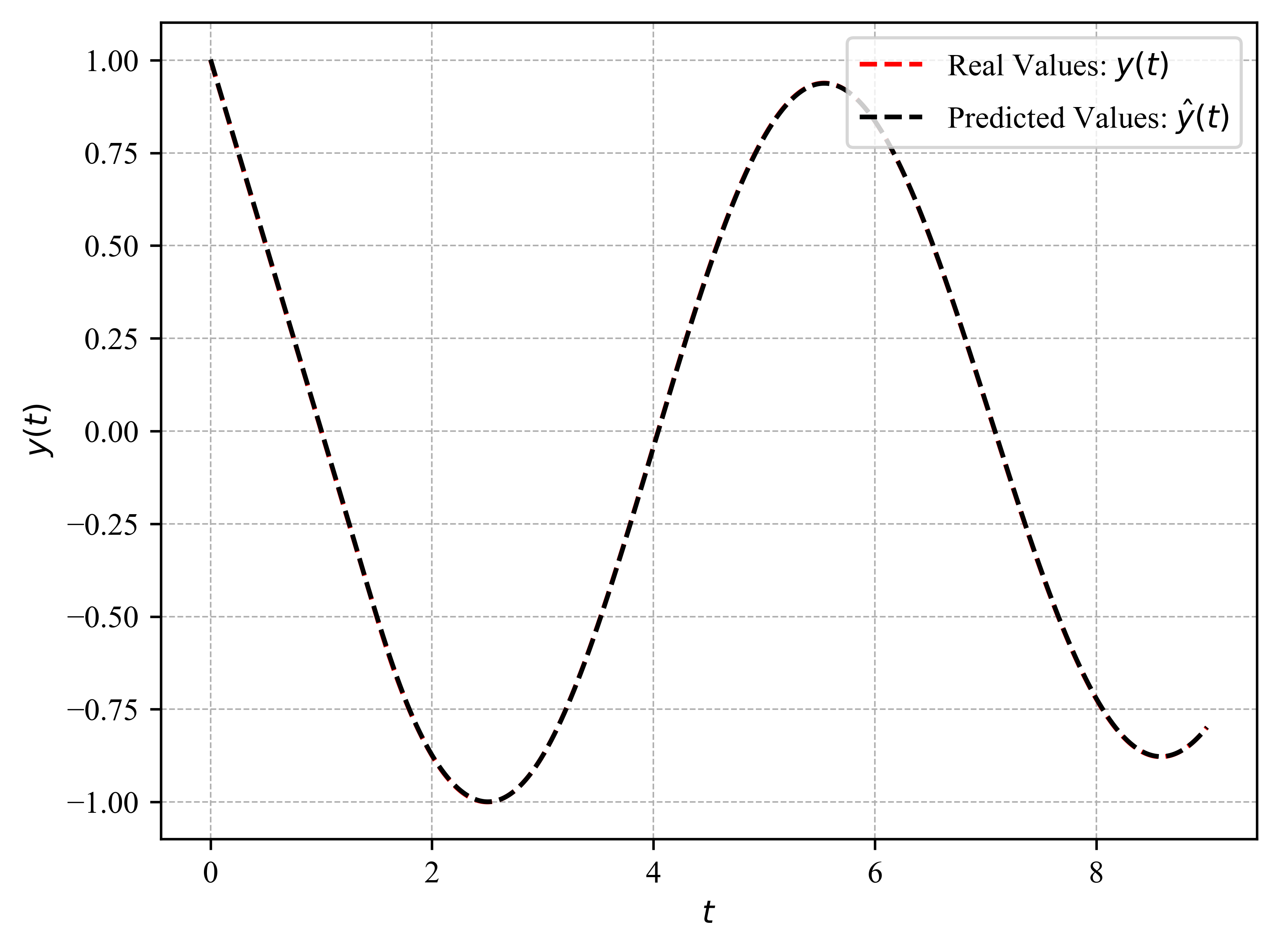}
		\caption{\footnotesize Trajectories of the final predicted vs. true trajectories for the delay differential equation with $\tau=1.5$ as shown in Eq. \eqref{equ17a}.}
		\label{fig:DDE_sol3}
	\end{minipage}\hfill
	\begin{minipage}{0.48\textwidth}
		\centering
		\includegraphics[height=5cm]{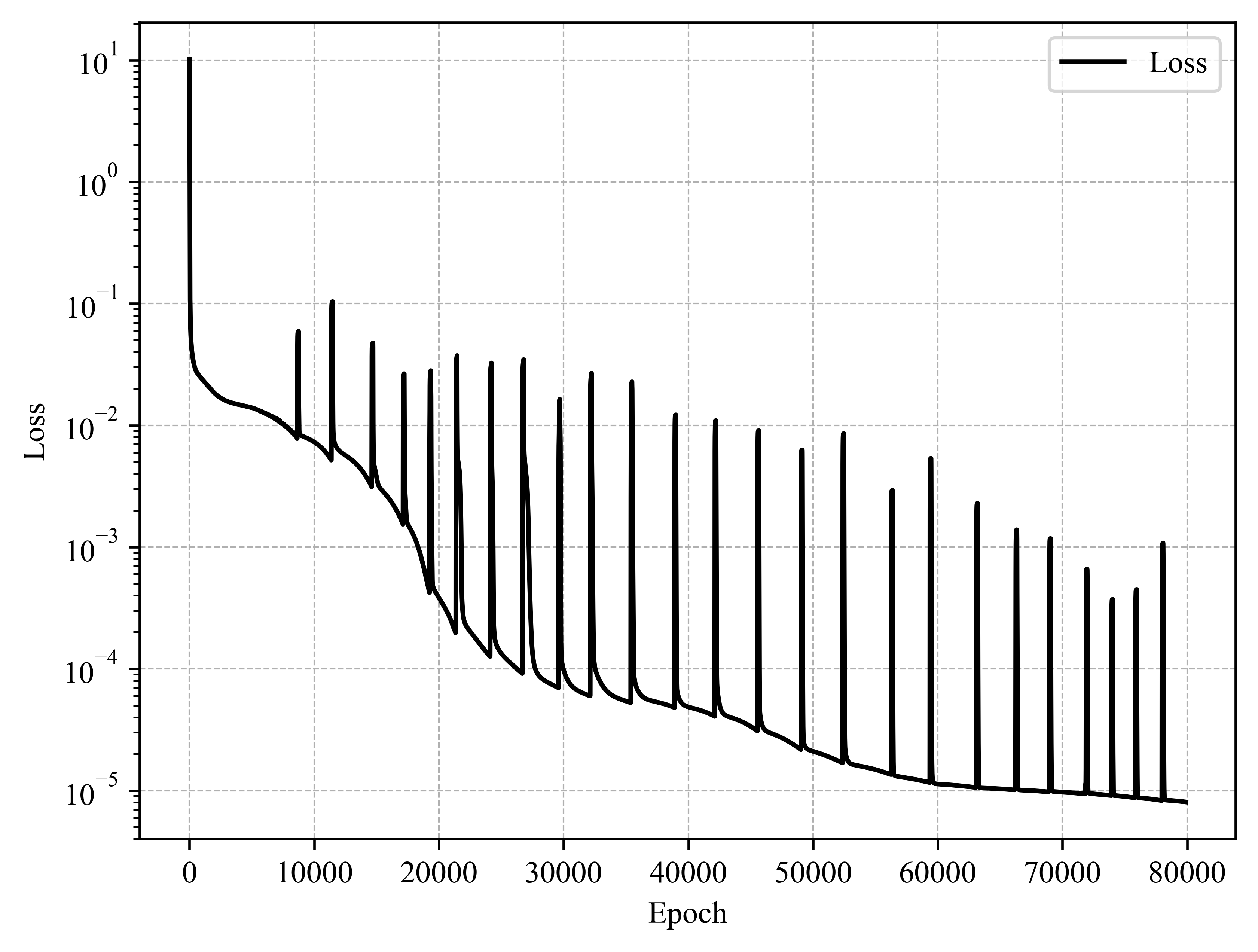}
		\caption{\footnotesize Loss convergence over iterations in NDDEs training when $\tau=1.5$.}
		\label{fig:DDE_loss3}
	\end{minipage}
\end{figure}

The comparison diagrams of the numerical solution after iteration and the true solution are shown, respectively, in Fig. \ref{fig:DDE_sol1}, Fig. \ref{fig:DDE_sol2}, and Fig. \ref{fig:DDE_sol3}. It is evident that NDDEs exhibit high accuracy in solving delay differential equations, as indicated by the overlap of the two solutions. The loss convergence over iterations in NDDEs training can be seen in Fig. \ref{fig:DDE_loss1}, Fig. \ref{fig:DDE_loss2}, and Fig. \ref{fig:DDE_loss3}.

\subsection{The Inverse Problem for Solving Delay Differential Equations}

\paragraph{Example 4.2.1} Consider the delay differential equation identical to the forward problem, as shown in Eq. \eqref{eq.DDE_IP_ex1}, but in this case, the delay parameter $\tau$ is unknown.
\begin{equation}
\begin{cases}
\begin{aligned}
        y'(t) = {}& -y(t-\tau),\qquad & t \geq 0\\
        y(t) = {}& 1, & t \leq 0
\end{aligned}
\end{cases}.
\label{eq.DDE_IP_ex1}
\end{equation}

Unlike the total loss for the forward problem shown in Eq. \eqref{eq.DDE_FP_NE_ex1_loss_total}, when solving the inverse problem, NDDEs incorporate the loss of the given data:
$$
    Loss_g = \dfrac{1}{N_g}\sum_{i=1}^{N_g}\left|y_{\mathrm{pred}}(t_i)-y_{\mathrm{exact}}(t_i)\right|^2,
$$
where $N_g$ denotes the number of given data points. Hence, the total loss in this inverse problem can be calculated as
$$
    Loss_{\mathrm{total}} = \omega_f \cdot Loss_f + \omega_i \cdot Loss_i + \omega_g \cdot Loss_g,
$$
The weights of the losses are determined by the adaptive weighting strategy described in Eq. \eqref{eq.DDE_IP_framework_weight}.

The training dataset consists of two parts: first, we randomly sample 5000 points $t_f$ on the solution interval $[0,10]$; then, we obtain known data points $\left(t_g,y_{\mathrm{exact}}(t_g)\right)$ on $t_g\in\{2,4,6,8,10\}$ through the analytical solution shown in Eq. \eqref{eq.DDE_FP_ex1_exact}.

We conduct three sets of numerical experiments, with the initial delay parameter in each experiment set to $\tau = 0$. The neural network consists of three hidden layers, with the number of neurons in each layer being 20, 40, and 20 respectively. The network parameters are optimized using the Adam optimizer, with each set of experiments undergoing 80,000 iterations. The experimental results are shown in the Tab. \ref{tab.DDE_IP_NE_ex1}, indicate that the relative error between the predicted and actual values of delay parameters for each set is less than 1\%, demonstrating that the NDDEs have successfully and precisely identified the unknown delay parameter.

\begin{table}[htbp]
  \centering
  \caption{Comparison of True and Predicted Time-Delay Parameters with Relative Errors}
    \begin{tabularx}{1\textwidth}{>{\centering\arraybackslash}X>{\centering\arraybackslash}X>{\centering\arraybackslash}X}
    \toprule
    True Value & Predicted Value & Relative Error \\
    \midrule
    0.5   & 0.4974976 & 0.50049\% \\
    1.0     & 0.9971283 & 0.28717\% \\
    1.5   & 1.4992174 & 0.05217\% \\
    \bottomrule
    \end{tabularx}
  \label{tab.DDE_IP_NE_ex1}
\end{table}%

\begin{figure}[!htp]
	\begin{minipage}{0.48\textwidth}
		\centering
		\includegraphics[height=5cm]{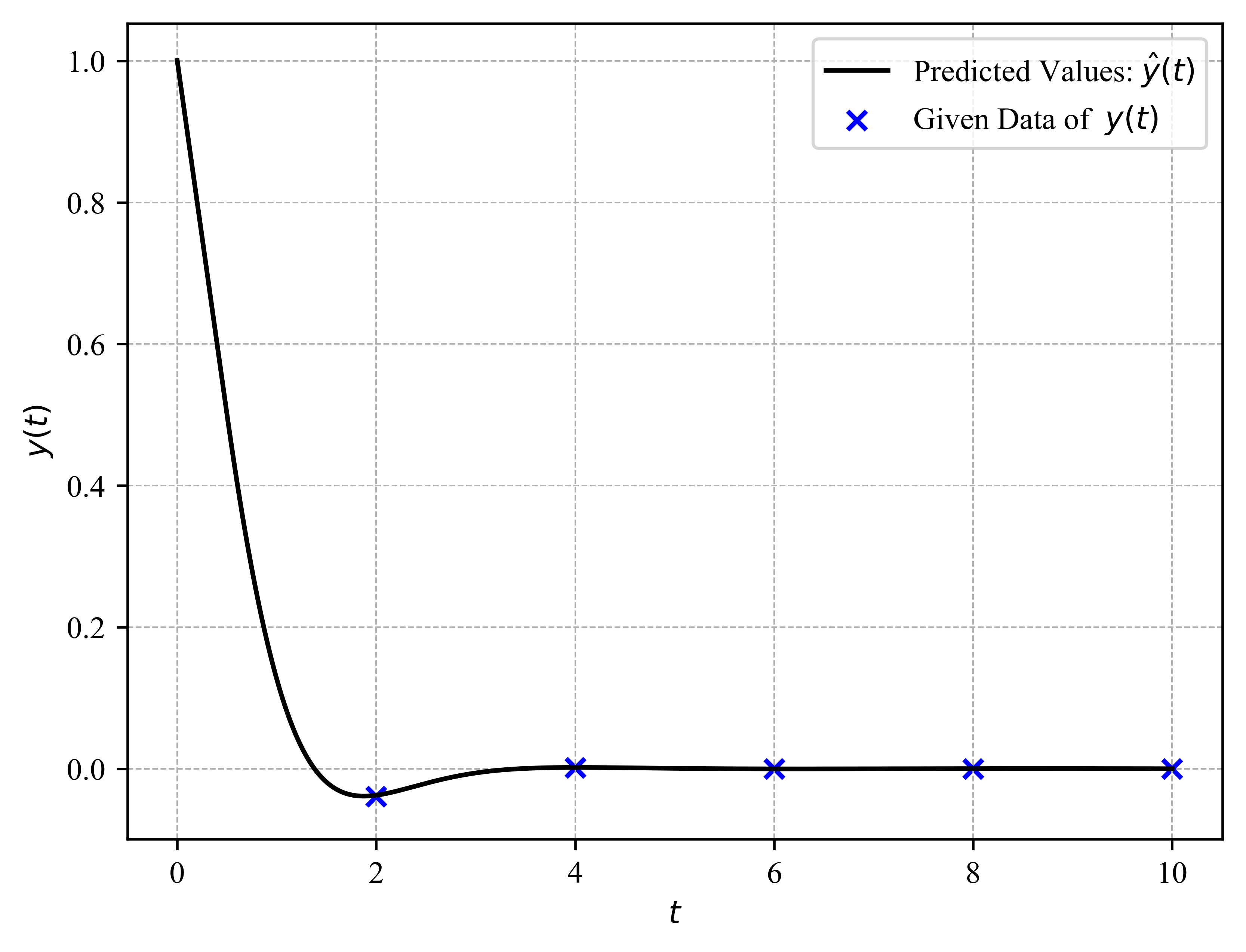}
            \caption{\footnotesize Known data and predicted values of $y(t)$ for the inverse problem of DDE: $\tau=0.5$.}
		\label{fig.DDE_lag_0.5}
	\end{minipage}\hfill
	\begin{minipage}{0.48\textwidth}
		\centering
		\includegraphics[height=5cm]{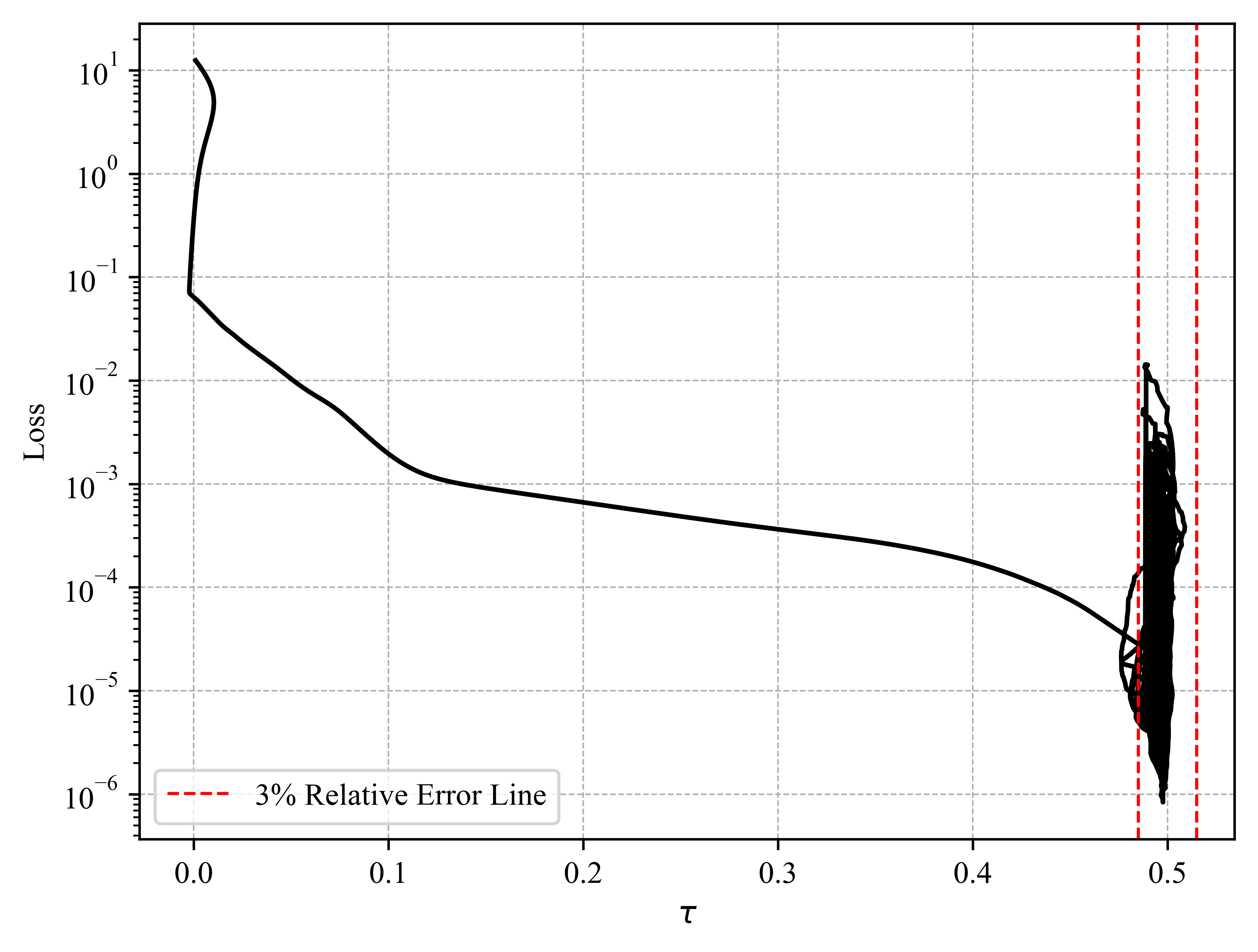}
		\caption{\footnotesize Delay parameter $\tau$ variation with loss in NDDEs training: $\tau=0.5$.}
		\label{fig.DDE_loss_lag_0.5}
	\end{minipage}

	\begin{minipage}{0.48\textwidth}
		\centering
		\includegraphics[height=5cm]{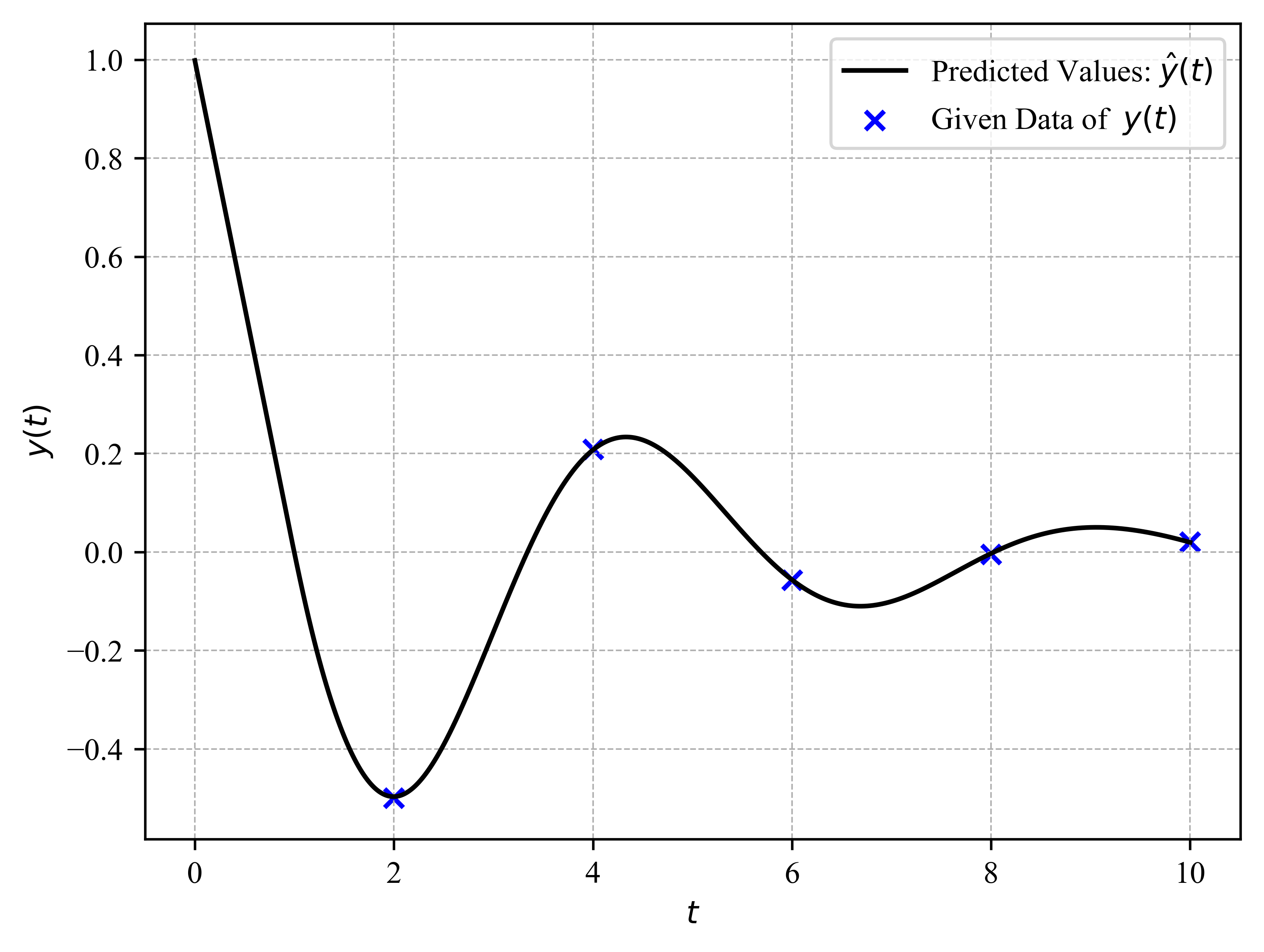}
            \caption{\footnotesize Known data and predicted values of $y(t)$ for the inverse problem of DDE: $\tau=1.0$.}
		\label{fig.DDE_lag_1.0}
	\end{minipage}\hfill
	\begin{minipage}{0.48\textwidth}
		\centering
		\includegraphics[height=5cm]{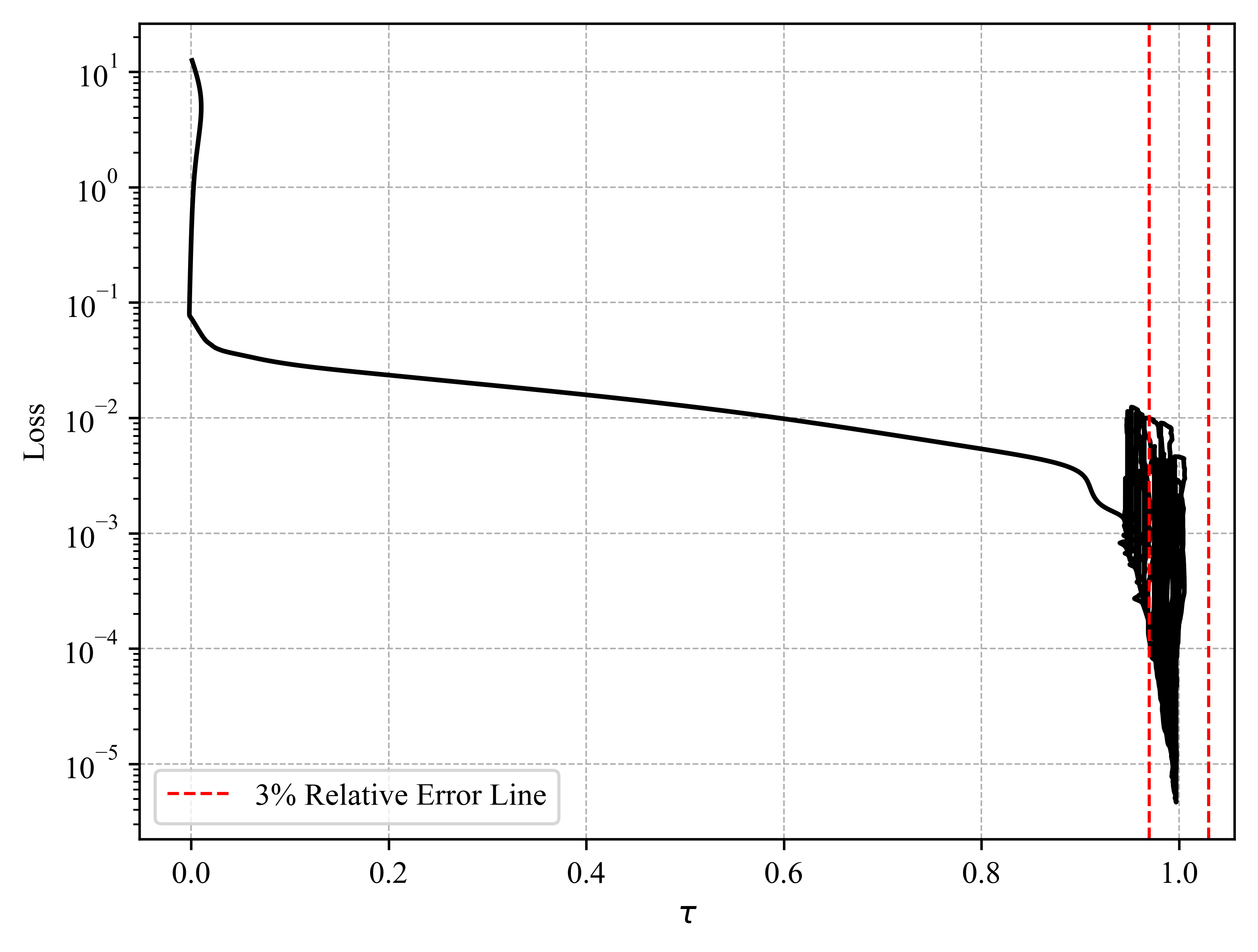}
		\caption{\footnotesize Delay parameter $\tau$ variation with loss in NDDEs training: $\tau=1.0$.}
		\label{fig.DDE_loss_lag_1.0}
	\end{minipage}

	\begin{minipage}{0.48\textwidth}
		\centering
		\includegraphics[height=5cm]{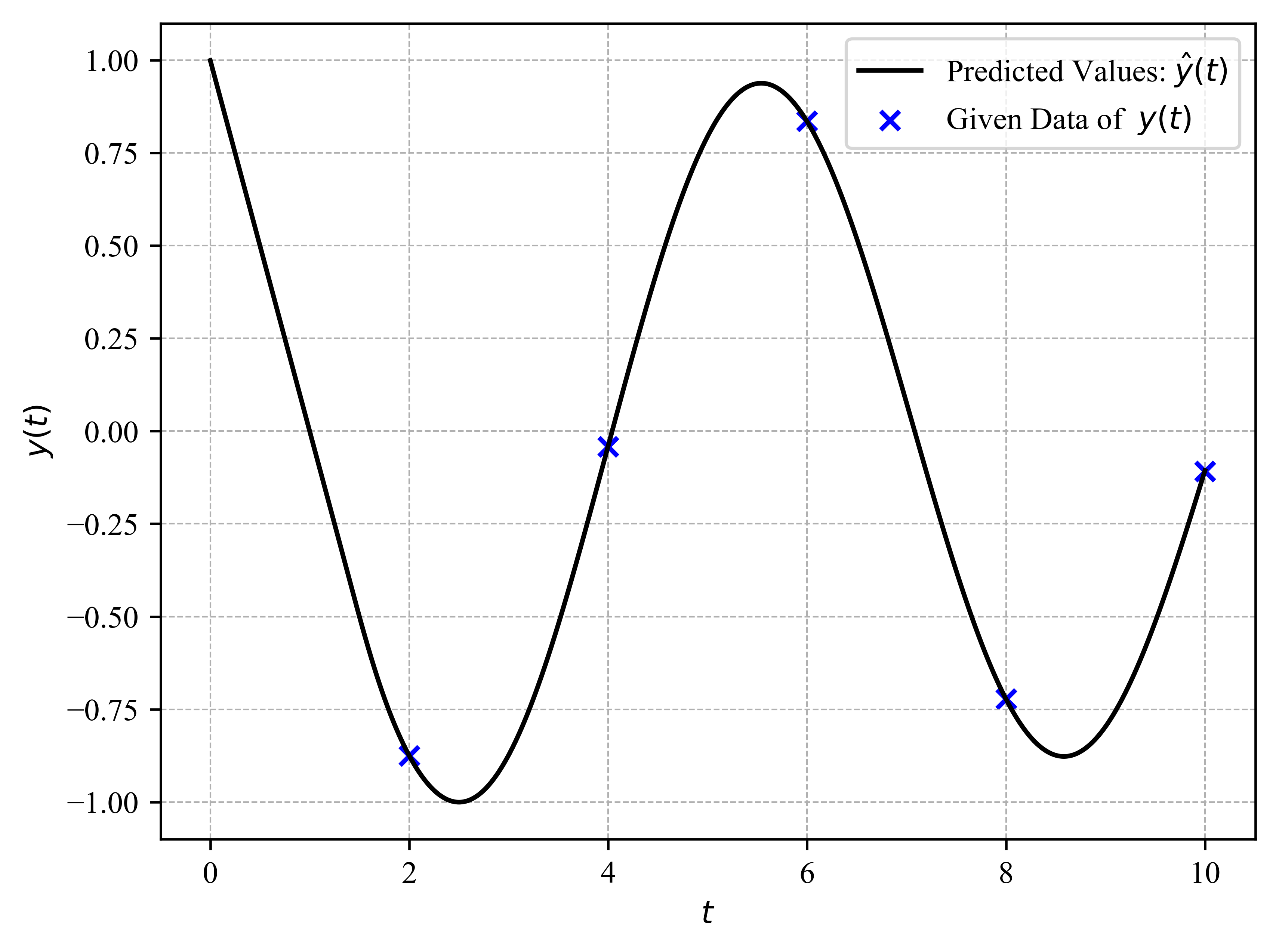}
            \caption{\footnotesize Known data and predicted values of $y(t)$ for the inverse problem of DDE: $\tau=1.5$.}
		\label{fig.DDE_lag_1.5}
	\end{minipage}\hfill
	\begin{minipage}{0.48\textwidth}
		\centering
		\includegraphics[height=5cm]{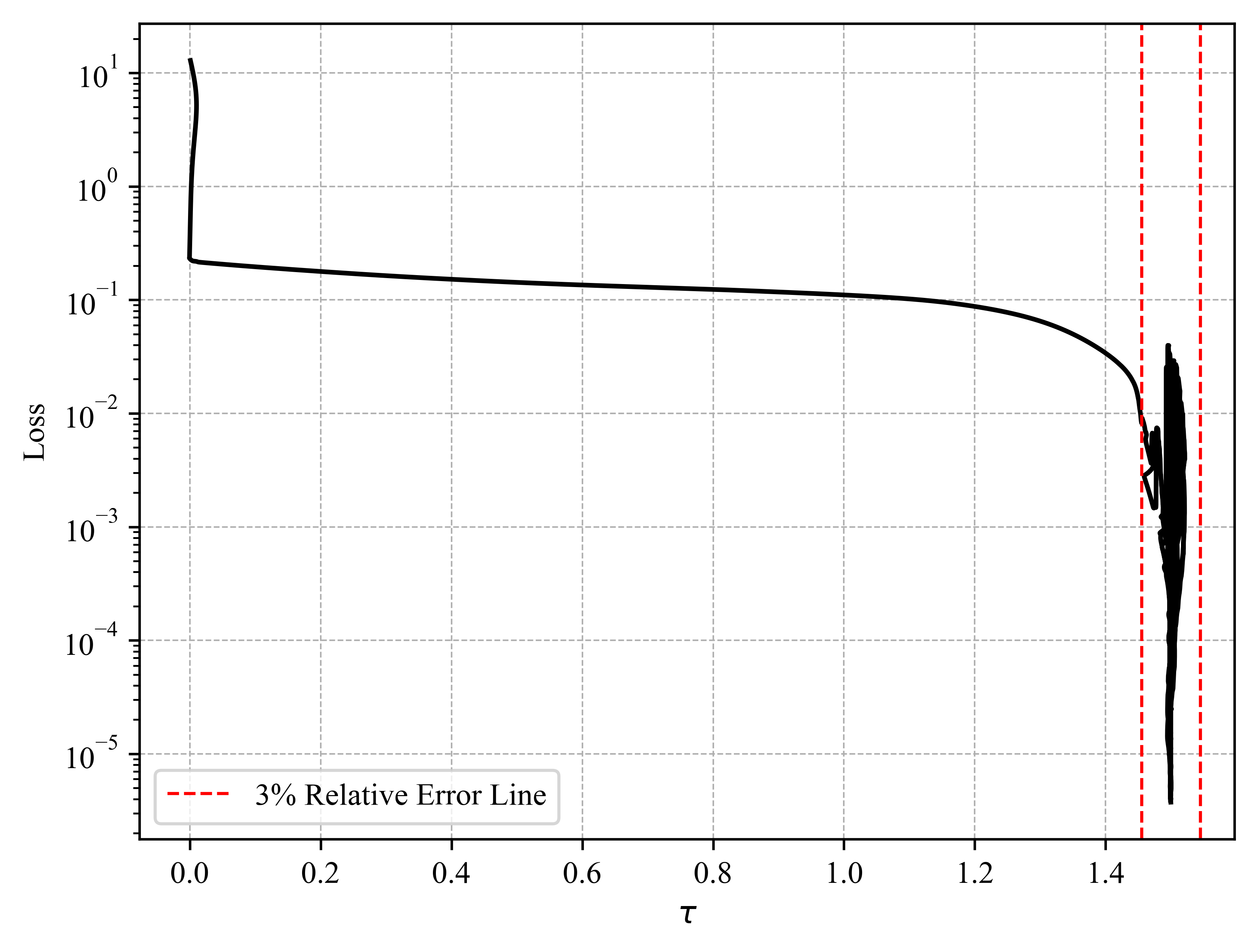}
		\caption{\footnotesize Delay parameter $\tau$ variation with loss in NDDEs training: $\tau=1.5$.}
		\label{fig.DDE_loss_lag_1.5}
	\end{minipage}
\end{figure}

For the actual values of the delay parameter $\tau$ being 0.5, 1.0, and 1.5, respectively, Fig. \ref{fig.DDE_lag_0.5}, \ref{fig.DDE_lag_1.0}, and \ref{fig.DDE_lag_1.5} illustrate the trajectories of the known data points and the predicted values. Fig. \ref{fig.DDE_loss_lag_0.5}, \ref{fig.DDE_loss_lag_1.0}, and \ref{fig.DDE_loss_lag_1.5} show the dynamic variations between $\tau$ and the loss values during the training process; the results indicate that as the loss gradually decreases, the predicted values of $\tau$ become increasingly close to their true values.

\paragraph{Example 4.2.2} Consider the classical logistic equation modified by Hutchinson\cite{hutchinson1948circular}:

\begin{equation}
    y'(t) = r y(t) \left(1 - \frac{y(t - \tau)}{K}\right).
\label{eq.DDE_IP_ex2}
\end{equation}

\noindent
Where the lag $\tau>0$ represents the maturation time of individuals in the population. The non-negative parameters $r$ and $K$ are known as the intrinsic growth rate and the environmental carrying capacity, respectively.

By setting $a=r$ and $b=\frac{r}{K}$, we can obtain the following delay differential equation:
\begin{equation}
    y'(t) = a \cdot y(t) - b \cdot y(t) \cdot y(t-\tau),
    \label{eq.DDE_IP_ex2_processed}
\end{equation}
assume that the initial condition of Eq. \eqref{eq.DDE_IP_ex2_processed} is
$$
    y(t) =  1,\quad  t \leq 0.
$$

In this experiment, the real delay parameter $\tau$ is set to 0.5, and the system parameter values for $r$ and $K$ are 0.6 and 100, respectively, resulting in $a=0.6$ and $b=0.006$. Known data points $\left(t_g, y_{\mathrm{exact}}(t_g)\right)$, with $t_g$ values of $\{1.00, 5.75, 10.50, 15.25, 20.00\}$, are derived using MATLAB's DDE23 solver. The initial delay parameter is set at $\tau = 0$ and the system parameters are set at $[a,b] = [0,0]$. The neural network is structured with three hidden layers that contain 20, 40, and 20 neurons, respectively. The network parameters are fine-tuned using the Adam optimizer in 80,000 iterations for the experimental run.

\begin{figure}[htbp]
    \centering
    \includegraphics[height=8cm]{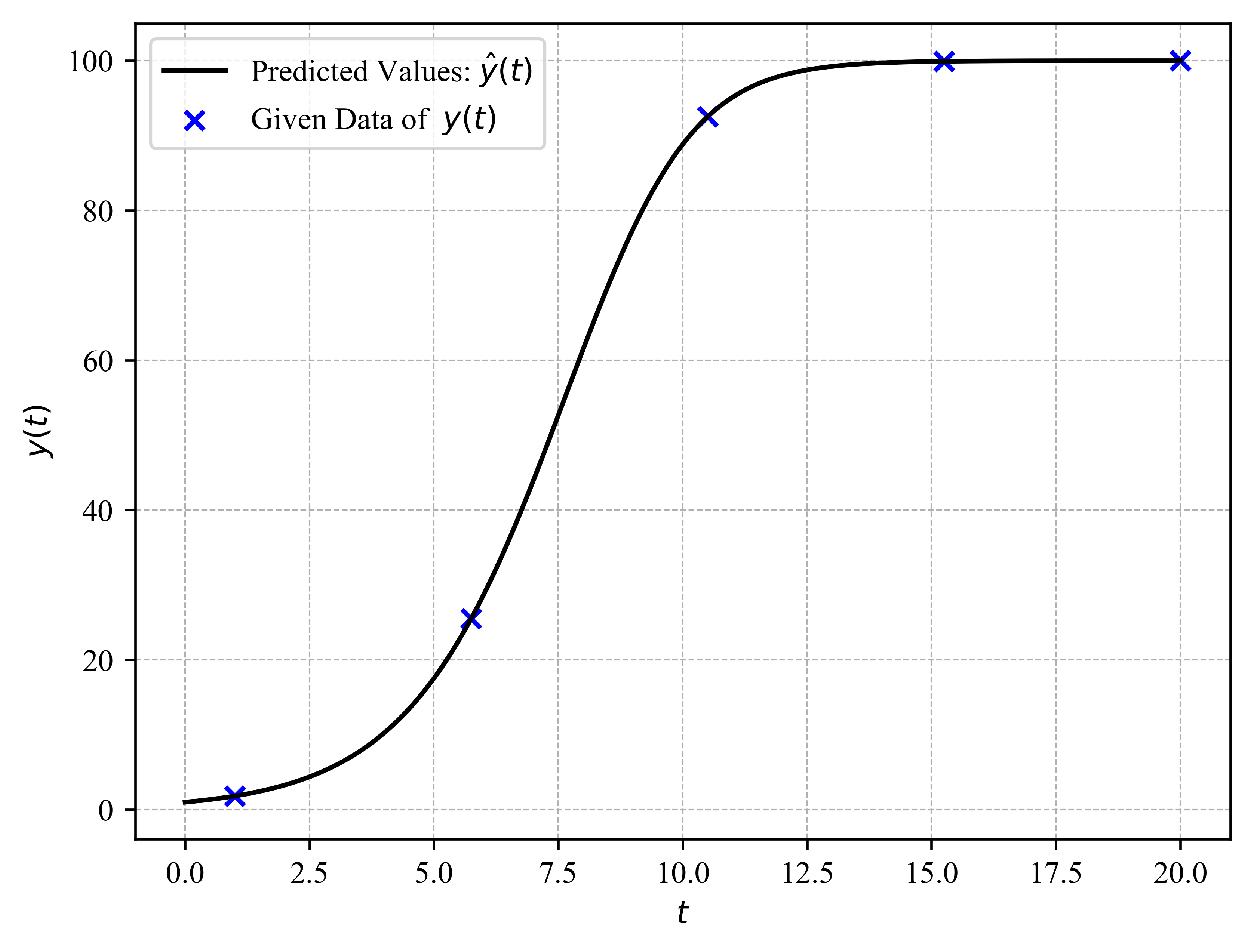}
    \caption{Given data and predicted values for the inverse problem of DDE: Example 2.}
    \label{fig.DDE_IP_ex2}
\end{figure}

The solution derived from the NDDEs is illustrated in Fig. \ref{fig.DDE_IP_ex2}. Tab. \ref{tab.DDE_IP_ex2} presents the experimental result, indicating that the relative error between the predicted and actual parameter values remains below 1\%.

\begin{table}[htbp]
  \centering
  \caption{Comparison of True and Predicted Parameters with Relative Errors}
    \begin{tabularx}{1\textwidth}{>{\centering\arraybackslash}X>{\centering\arraybackslash}X>{\centering\arraybackslash}X>{\centering\arraybackslash}X}
    \toprule
    Parameter & True Value & Predicted Value & Relative Error \\
    \midrule
    $\tau$    & 0.500   & 0.50195070 & 0.3901\% \\
    $a$       & 0.600   & 0.59912218 & 0.1463\% \\
    $b$       & 0.006   & 0.00599196 & 0.1340\% \\
    \bottomrule
    \end{tabularx}
  \label{tab.DDE_IP_ex2}
\end{table}%

\subsection{The Forward Problem for Solving the System of Delay Differential Equations}
Consider the system of delay differential equations with  with quadruple delays shown in Eq. \eqref{eq.DDEs_FP_NE_ex1} on the solution interval $[0,1]$:
\begin{equation}
    \begin{cases}
    \begin{aligned}
    y_1'(t) ={} & y_1(t-0.2)\\
    y_2'(t) ={} & y_1(t-0.3) \cdot y_2(t-0.4)\\
    y_3'(t) ={} & y_2(t-0.5)
    \end{aligned}
    \end{cases}.
    \label{eq.DDEs_FP_NE_ex1}
\end{equation}
The initial conditions for Eq. \eqref{eq.DDEs_FP_NE_ex1} are given by
\begin{equation}
    y_j(t) = 1,~\mathrm{for}\; t \in [-1, 0],\; j=1,2,3.
\label{eq.DDEs_FP_NE_ex1_IC}
\end{equation}

Referring to the loss construction methods presented in Sec. \ref{subsec.DDEs_Framework_FP}, we can obtain the loss for each equation within the system of delay differential equations:
\begin{align}
    Loss_{f,1} = {}& \dfrac{1}{N_f}\sum_{k=1}^{N_f}\left|\frac{\mathrm{d}}{\mathrm{d}t}y^{\mathrm{pred}}_1(t_k) - y^{\mathrm{pred}}_1(t_k-0.2)\right|^2,
\label{eq.DDEs_FP_NE_ex1_Loss_f1}\\
    Loss_{f,2} = {}& \dfrac{1}{N_f}\sum_{k=1}^{N_f}\left|\frac{\mathrm{d}}{\mathrm{d}t}y^{\mathrm{pred}}_2(t_k) - y^{\mathrm{pred}}_1(t_k-0.3) \cdot y^{\mathrm{pred}}_2(t_k-0.4) \right|^2,
\label{eq.DDEs_FP_NE_ex1_Loss_f2}\\
    Loss_{f,3} = {}& \dfrac{1}{N_f}\sum_{k=1}^{N_f}\left|\frac{\mathrm{d}}{\mathrm{d}t}y^{\mathrm{pred}}_3(t_k) - y^{\mathrm{pred}}_2(t_k-0.5)\right|^2.
\label{eq.DDEs_FP_NE_ex1_Loss_f3}
\end{align}
The loss of the initial condition for each equation is calculated by
\begin{equation}
    Loss_{i,j} = \left|y^{\mathrm{pred}}_j(0)-1\right|^2.
\label{eq.DDEs_FP_ex1_Loss_i}
\end{equation}
The total loss in the NDDEs framework is the weighted sum of all the associated losses, which is expressed in Eq. \eqref{eq.DDEs_FP_framework_loss_total}:
\begin{equation}
    Loss_{\mathrm{total}} = \sum_{j=1}^{3} \left( \omega_{f,j} \cdot Loss_{f,j} + \omega_{i,j} \cdot Loss_{i,j} \right).
\label{eq.DDEs_FP_ex1_Loss_total}
\end{equation}
The weights of the losses are determined by the adaptive weighting strategy described in Eq. \eqref{eq.DDEs_FP_weight_f} and Eq. \eqref{eq.DDEs_FP_weight_i}.

\begin{figure}[htbp]
    \centering
    \includegraphics[height=8cm]{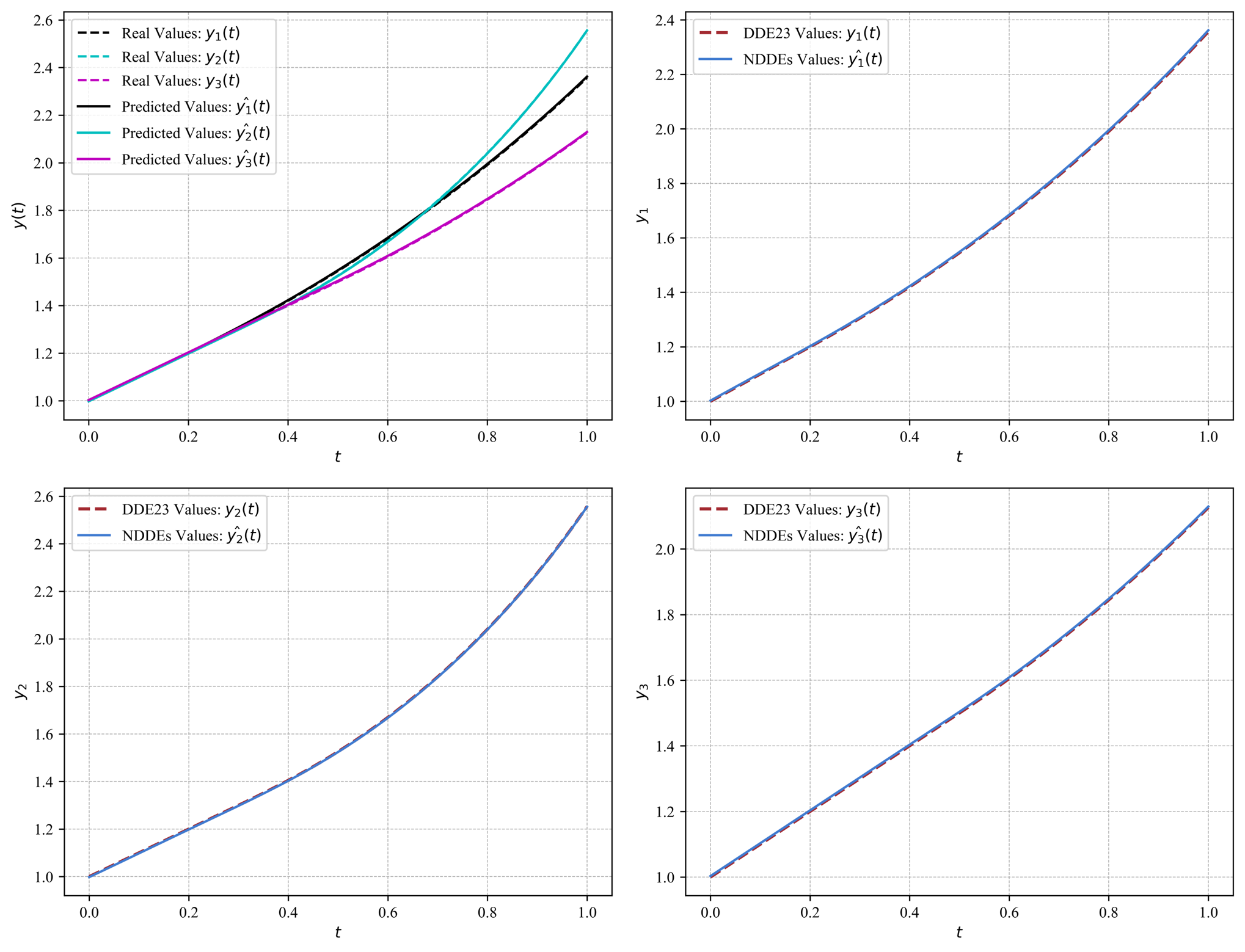}
    \caption{Solutions of the system of DDEs using NDDEs and DDE23.}
    \label{fig.DDEs_FP_sol}
\end{figure}

Our designed neural network consists of three hidden layers, containing $20$, $40$, and $20$ neurons respectively. To compare with traditional numerical methods, we also use MATLAB's DDE23 function to solve this set of equations. The DDE23 function, which integrates using an explicit Runge-Kutta (2,3) pair and an ODE23 interpolator, is capable of handling situations where the lag time exceeds the step size. The comparison diagrams of the numerical solution after 80,000 iterations and the solution from DDE23 are shown, respectively, in Fig. \ref{fig.DDEs_FP_sol}. Tab. \ref{tab.DDEs_FP_ex1_L2} indicates that the relative error of the numerical solution obtained by NDDEs, compared to the result of DDE23, is within 1\%.

\begin{table}[htbp]
  \centering
  \caption{The Relative Error between the NDDEs Solution and DDE23 Solution}
    \begin{tabularx}{1\textwidth}{>{\centering\arraybackslash}X>{\centering\arraybackslash}X>{\centering\arraybackslash}X>{\centering\arraybackslash}X}
    \toprule
    $y(t)$  & \multicolumn{1}{c}{$y_1$} & \multicolumn{1}{c}{$y_2$} & \multicolumn{1}{c}{$y_3$} \\
    \midrule
    RE & 0.21402\% & 0.10952\% & 0.21787\% \\
    \bottomrule
    \end{tabularx}
  \label{tab.DDEs_FP_ex1_L2}
\end{table}

\subsection{The Inverse Problem for Solving the System of Delay Differential Equations}
\label{subsec.DDEs_IP}
\paragraph{Example 4.4.1} Consider the system of delay differential equations, characterized by quadruple unknown delays, as presented in Eq. \eqref{eq.DDEs_IP_NE_ex1}:
\label{example.1}

\begin{equation}
    \begin{cases}
    \begin{aligned}
    y_1'(t) = {} & y_1(t-\tau_1)\\
    y_2'(t) = {} & y_1(t-\tau_2) \cdot y_2(t-\tau_3)\\
    y_3'(t) = {} & y_2(t-\tau_4)
    \end{aligned}
    \end{cases}
\label{eq.DDEs_IP_NE_ex1}
\end{equation}
The initial conditions for are given by
\begin{equation}
    y_j(t) = 1,~\mathrm{for}\; t \in [-1, 0],\; j=1:3.
\label{eq.DDEs_IP_NE_ex1_IC}
\end{equation}
Within the framework of NDDEs for inverse problems, the definitions of the associated losses for the system of delay differential equations are given in Eq. \eqref{eq.DDEs_FP_framework_loss_f}, \eqref{eq.DDEs_FP_framework_loss_i}, and \eqref{eq.DDEs_IP_framework_loss_g}.

The training dataset consists of two parts: first, we randomly sample 5000 points $t_f$ on the solution interval $[0,1]$; then, we obtain known data points $\left(t_g,y^{\mathrm{exact}}_j(t_g)\right)$ on $t_g\in\{0.2,0.4,0.6,0.8,1.0\}$ through MATLAB's DDE23 solver.

We conduct three sets of numerical experiments, with the initial delay parameters in each experiment set to $[\tau_1, \tau_2, \tau_3, \tau_4] = [0,0,0,0]$. The neural network consists of three hidden layers, with the number of neurons in each layer being 20, 40, and 20, respectively. The network parameters are optimized using the Adam optimizer, with each set of experiments undergoing 160,000 iterations. The experimental results, as shown in Tab. \ref{tab.DDEs_IP_NE_ex1}, indicate that the relative error between the predicted and actual values of delay parameters for each set is less than 3\%, demonstrating that the NDDEs have successfully identified the unknown delay parameters.

\begin{table}[htbp]
	\centering
 \caption{Comparison of True and Predicted Time-Delay Parameters with Relative Errors}
	\begin{tabularx}{1\textwidth}{>{\centering\arraybackslash}X>{\centering\arraybackslash}X>{\centering\arraybackslash}X>{\centering\arraybackslash}X>{\centering\arraybackslash}X>{\centering\arraybackslash}X>{\centering\arraybackslash}X>{\centering\arraybackslash}X>{\centering\arraybackslash}X>{\centering\arraybackslash}X}
		\toprule
		& \multicolumn{3}{c}{Group 1} & \multicolumn{3}{c}{Group 2} & \multicolumn{3}{c}{Group 3} \\
		\midrule
		& TV & PV & RE & TV & PV & RE & TV & PV & RE \\
		\midrule
        $\tau_1$   & 0.1 & 0.0992 & 0.83\% & 0.2 & 0.1942 & 2.90\% & 0.6 & 0.5896 & 1.73\% \\
        $\tau_2$   & 0.2 & 0.2036 & 1.79\% & 0.3 & 0.2946 & 1.79\% & 0.5 & 0.4979 & 0.41\% \\
        $\tau_3$   & 0.3 & 0.2933 & 2.23\% & 0.4 & 0.4114 & 2.85\% & 0.4 & 0.4017 & 0.43\% \\
        $\tau_4$   & 0.4 & 0.3988 & 0.30\% & 0.5 & 0.5027 & 0.53\% & 0.3 & 0.3023 & 0.75\% \\
		\bottomrule
	\end{tabularx}

	\begin{tablenotes}
	\small
	\item Note: TV: True Value, PV: Predicted Value, RE: Relative Error.
	\end{tablenotes}
	\label{tab.DDEs_IP_NE_ex1}
	\end{table}

\begin{figure}[htbp]
    \centering
    \includegraphics[height=3.8cm]{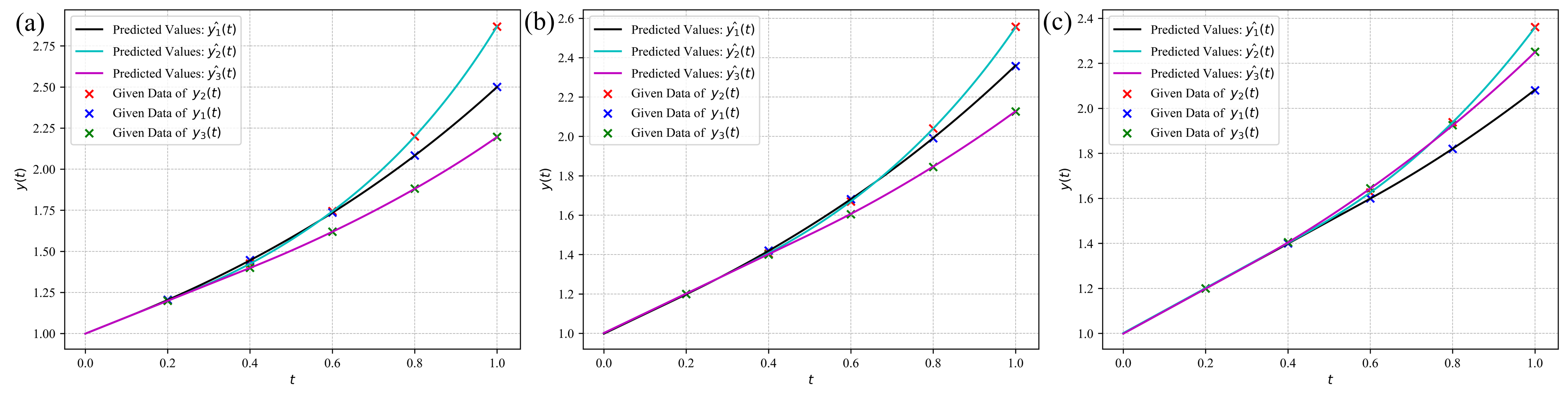}
    \caption{Given data and predicted values of $y_j(t)$ for the inverse problem of the system of DDEs. (a): Group 1, (b): Group 2, (c): Group 3.}
    \label{fig.DDEs_IP_ex_2}
\end{figure}

\begin{figure}[htbp]
    \centering
    \includegraphics[height=8.5cm]{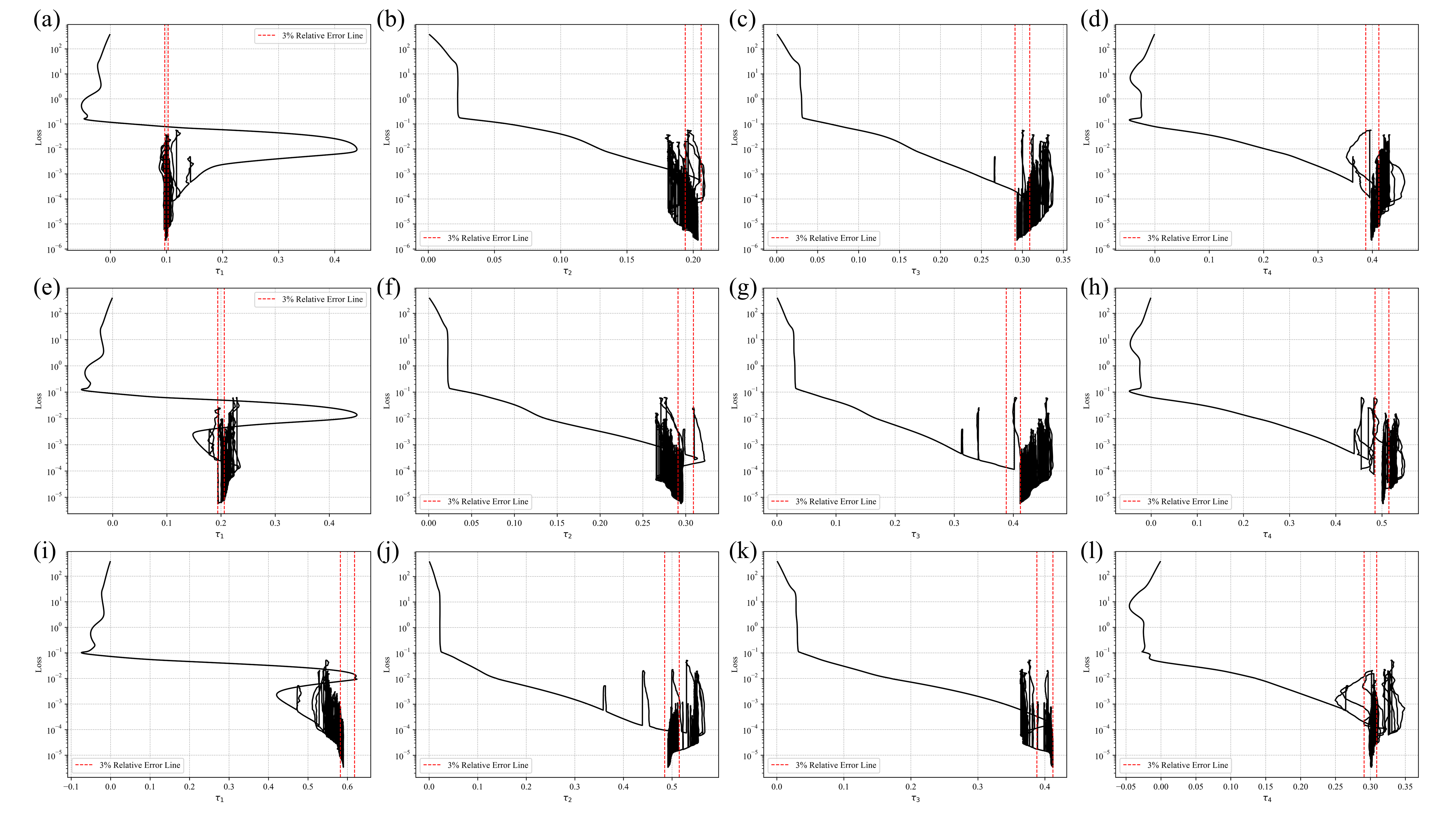}
    \caption{Variation of delay parameter $\tau_i$ with loss in NDDEs training. (a-d): Group 1, (e-h): Group 2, (i-l): Group 3.}
    \label{fig.DDEs_IP_NE_ex1_2_loss}
\end{figure}

Fig. \ref{fig.DDEs_IP_ex_2} plotted the known data points and the trajectory of predicted values respectively, and Fig. \ref{fig.DDEs_IP_NE_ex1_2_loss} revealed the dynamic changes between the delay parameters and the loss value during training; the results indicated that as the loss gradually decreased, the predicted value of the delay parameters increasingly approached its true value.

\paragraph{Example 4.4.2} Consider the enhanced SIR model proposed by Rihan et al. \cite{rihan2021delay}: 
\begin{equation}
    \begin{cases}
        \begin{aligned}\dot{S}(t) &= -\beta \cdot S(t) \cdot I(t-\tau_2) + \gamma \cdot I(t-\tau_1), \quad t \geq 0, \\\dot{I}(t) &= \beta \cdot S(t) \cdot I(t-\tau_2) - \alpha \cdot  I(t), \quad t \geq 0, \\\dot{R}(t) &= \alpha \cdot I(t) - \gamma \cdot I(t-\tau_1), \quad t \geq 0.\end{aligned}
    \end{cases}
    \label{Eq.IP_DDEs_SIR}
\end{equation}
In this model, $S(t)$ quantifies the susceptible segment of the population, $I(t)$ refers to those infected, and $R(t)$ indicates individuals who have been removed, either through recovery or otherwise. The term $\beta$ denotes the rate of infection per unit time, while $\alpha$ represents the proportion of individuals transitioning out of the infectious state (removal rate), and $\gamma$ captures the recovery rate, signifying the transition frequency from infected to removed status. Two delay parameters, $\tau_1$ and $\tau_2$, are concurrently integrated into the model. Here, $\tau_1$ indicates the period after which individuals with immunity become once again susceptible and $\tau_2$ delineates the latency period from initial exposure to the manifestation of infection.

The initial conditions for the system of delay differential equations referred to in Eq. \eqref{Eq.IP_DDEs_SIR} are as follows
$$
    [S(0), I(0), R(0)] = [5, 1, 0],
$$
and the system parameters are set to
$$
    [\alpha, \beta, \gamma] = [0.7, 0.3, 0.1].
$$

The training dataset is composed of two segments: initially, 5000 points denoted by $t_f$ are randomly sampled across the solution interval $[0,10]$. Subsequently, the known data points $\left(t_g, y^{\mathrm{exact}}_j(t_g)\right)$, where $t_g$ is within the set $\{2, 4, 6, 8, 10\}$, are obtained using MATLAB's DDE23 solver. 

\begin{figure}[htbp]
    \centering
    \includegraphics[height=7cm]{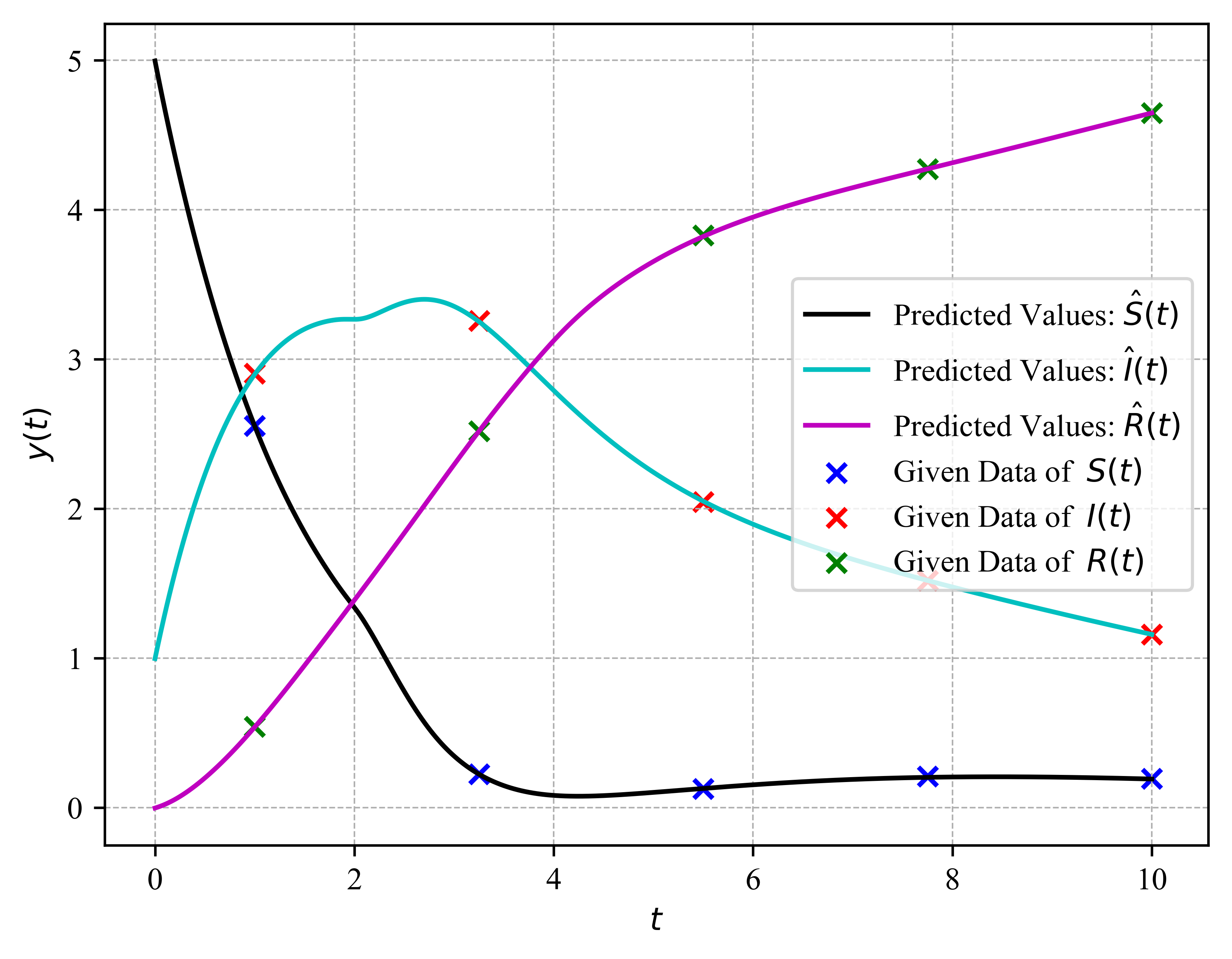}
    \caption{Given data and predicted values for the inverse problem of the system of DDEs: Example 4.}
    \label{fig.DDEs_IP_SIR}
\end{figure}

The solutions obtained from the NDDEs following 80,000 iterations are depicted in Figure \ref{fig.DDEs_IP_SIR}. Table \ref{tab.DDEs_IP_SIR} displays the experimental outcomes, illustrating that the relative error between the forecasted and true values of the delay parameters for each experimental set remains under 2\%.

\begin{table}[htbp]
  \centering
  \caption{Comparison of True and Predicted Time-Delay Parameters with Relative Errors}
    \begin{tabularx}{1\textwidth}{>{\centering\arraybackslash}X>{\centering\arraybackslash}X>{\centering\arraybackslash}X>{\centering\arraybackslash}X}
    \toprule
     & True Value & Predicted Value & Relative Error \\
    \midrule
    $\tau_1$    & 4.0   & 3.9536 & 1.1590\% \\
    $\tau_2$    & 2.0   & 1.9915 & 0.4249\% \\
    \bottomrule
    \end{tabularx}
  \label{tab.DDEs_IP_SIR}
\end{table}%

%================== Conclusion ====================%
\section{Conclusion}
In this article, we propose a framework for Neural Delay Differential Equations (NDDEs), which is used to solve the forward and inverse problems of delay differential equations. NDDEs use deep neural network training models to calculate derivatives via automatic differentiation, and give full consideration to initial value conditions, control equations, and known data. NDDEs can perform calculations directly across the entire solution region, thus avoiding the complexities of discretization. Since the neural network itself provides an approximate analytical representation of the solution, it also avoids additional computational costs and discretization errors. Therefore, using NDDEs can provide continuous and differentiable solutions for delay differential equations.

Through numerous numerical experiments, we have verified the outstanding performance of NDDEs in solving forward and inverse problems of delay differential equations with single or multiple delays. In solving forward problems, NDDEs can accurately solve the delay differential equation or the systems of delay differential equations with multiple delays, with results comparable to traditional numerical analysis methods; in solving inverse problems, NDDEs can effectively deduce delay parameters as well as multiple system parameters in the system of delay differential equations, with an accuracy very close to the true values.

%================== authorship contribution statement ====================%
\section*{Authorship Contribution Statement}
\textbf{Housen Wang}: Background Research, Conceptualization and Design, Methodology, Coding, Data Curation, Validation, Visualization, Writing – original draft, Project Administration.
\textbf{Yuxing Chen}: Background Research, Methodology, Validation, Writing – original draft.
\textbf{Sirong Cao}:  Background Research, Visualization, Writing - Original draft.
\textbf{Xiaoli Wang}: Background Research, Writing - Original draft.
\textbf{Qiang Liu}: Conceptualization, Writing - Review \& Editing, Supervision, and Funding acquisition.

\section*{Acknowledgement}
The authors would like to thank the Prof. Ming Mei for providing us with important references and examples.

%==============================================%
% \clearpage
\bibliographystyle{elsarticle-num}
\bibliography{BibTex}
\end{document}